\documentclass[letterpaper,journal]{IEEEtran}
\usepackage{amsmath,amsfonts,amssymb}
\usepackage{algorithm}
\usepackage{algpseudocode}
\usepackage{array}
\usepackage{booktabs}
\usepackage{capt-of}
\usepackage{float}
\usepackage{placeins}
\usepackage{graphicx}
\usepackage{stfloats}
\usepackage{textcomp}
\usepackage{url}
\usepackage{cite}
\usepackage{pifont}
\usepackage[table]{xcolor}
\definecolor{generalistblue}{RGB}{224,236,246}
\definecolor{specialistgreen}{RGB}{229,240,220}
\definecolor{stageoneblue}{RGB}{242,247,251}
\definecolor{fullgreen}{RGB}{242,248,238}
\makeatletter
\let\ExtremeRGMTOriginalMakeCaption\@makecaption
\long\def\@makecaption#1#2{%
\ifx\@captype\@IEEEtablestring
\footnotesize\bgroup\par\@IEEEtabletopskipstrut\noindent
{\normalfont\footnotesize #1: #2}\par\addvspace{0.5\baselineskip}\egroup%
\@IEEEtablecaptionsepspace
\else
\ExtremeRGMTOriginalMakeCaption{#1}{#2}%
\fi}
\makeatother
\graphicspath{{./}}
\begin{document}

\title{Extreme-RGMT: Continual Learning of Highly Dynamic Skills for Robust Generalist Humanoid Control}

\author{Yubiao Ma\textsuperscript{1,2,*}, Han Yu\textsuperscript{2,*}, Kai Guo$^{3}$, Changtai Lv\textsuperscript{2},
Zhengquan Mao\textsuperscript{2}, Boyang Xing\textsuperscript{2}, Xuemei Ren\textsuperscript{1},
and Dongdong Zheng\textsuperscript{1,2}%
\thanks{\textsuperscript{1}School of Automation, Beijing Institute of Technology, Beijing 100081, China.
Emails: ybma@bit.edu.cn; xmren@bit.edu.cn; ddzheng@bit.edu.cn.}%
\thanks{\textsuperscript{2}Humanoid Robotics (Shanghai) Co., Ltd., Shanghai 201203, China.
Emails: yuhan@openloong.net; lvchangtai@openloong.net; maozhengquan@openloong.net; xby@openloong.net.}%
\thanks{\textsuperscript{3}School of Mechanical Engineering, Shandong University, Jinan 250061, China. Email: {\small\texttt{kaiguo@sdu.edu.cn}}}%
\thanks{\textsuperscript{*}Equal contribution.}%
\thanks{Corresponding authors: Dongdong Zheng and Kai Guo.}}

\maketitle
\thispagestyle{empty}

\begin{abstract}
Humans can progressively acquire highly dynamic motor skills while preserving reliable everyday motor abilities. In contrast, existing humanoid controllers face a trade-off between generalist and specialist capabilities: generalist motion tracking policies struggle to reliably execute rare highly dynamic motions, whereas specialist training can degrade previously acquired behaviors. We introduce Extreme-RGMT, a two-stage continual learning framework for robust generalist humanoid control. The method first learns a generalist motion-tracking base policy from diverse multi-source motion data, then employs an asymmetric skill acquisition and capability consolidation mechanism to constrain policy drift on mastered motions while emphasizing difficult dynamic segments. To address the scarcity of highly dynamic motions, their high failure rates, and the resulting shortage of informative samples, Extreme-RGMT combines difficulty-aware sampling with advantage-prioritized trajectory resampling to emphasize critical segments. Experiments show that Extreme-RGMT achieves state-of-the-art generalist whole-body motion-tracking performance, including substantially improved completion of challenging highly dynamic motions. The resulting controller directly executes diverse unseen highly dynamic motions under fixed references and online inertial motion-capture inputs, advancing generalist whole-body motion-tracking controllers toward highly dynamic motor capabilities at the human-expert level.
{Project page:} \url{https://zeonsunlightyu.github.io/Extreme-RGMT.github.io/}

\end{abstract}

\begin{IEEEkeywords}
Humanoid robots, continual robot learning, motion tracking.
\end{IEEEkeywords}

\section{Introduction}

A central goal of humanoid robot control is to enable robots to progressively acquire whole-body motor capabilities approaching those of human experts. Human experts can perform complex motions that are highly dynamic, contact-rich, and strongly coordinated, and these capabilities are usually built upon stable fundamental motor skills. Through long-term practice, humans can expand their motor boundaries while retaining previously acquired abilities~\cite{Ericsson1993DeliberatePractice,Adolph2012LearningToWalk,BrashersKrug1996Consolidation}. This progressive development from fundamental motor abilities to expert skills provides a natural inspiration for humanoid robot learning: an ideal humanoid controller should possess broad and stable generalist motor competence, and should be able to acquire more challenging dynamic skills on top of it.

\begin{figure}[!t]
    \centering
    \includegraphics[width=0.95\columnwidth,keepaspectratio]{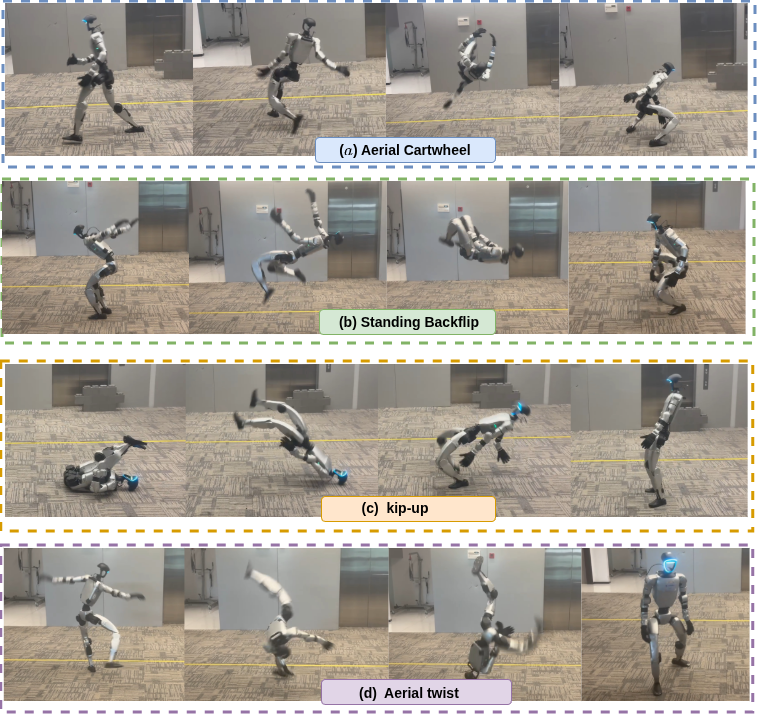}
    \caption{Qualitative real-world results. Representative physical rollouts of Extreme-RGMT on highly dynamic motions, including aerial cartwheel, standing backflip, kip-up, and aerial twist. The robot executes rapid aerial adjustments, contact transitions, and landing recovery on the Unitree G1 humanoid.}
    \label{fig:replay}
\end{figure}

However, existing generalist motion tracking and online teleoperation methods are still far from this goal. Prior methods can train a unified controller from large-scale motion libraries and support basic locomotion, simple pose following, and regular whole-body motions, demonstrating the potential of humanoid robots to respond to human motion inputs in real time~\cite{Luo2025SONIC,Ma2026RAGMT,GMT_2025,UniTracker_2025,KungfuBot2_2025,Li2025BFMZero,Li2025CLONE,Pan2025AMS,Serifi2024VMP,TWIST2_2025,Ben2025HOMIE,Cheng2024Expressive}. Nevertheless, these capabilities mainly cover relatively smooth motion regimes. For expert-level highly dynamic motions, existing controllers still struggle to achieve reliable execution. A key reason is that the effective control requirements of highly dynamic motions are concentrated in a small number of short temporal segments, such as aerial posture adjustment and landing recovery. Once the policy fails in these segments, the resulting rollout contains few informative successful transitions. In large-scale generalist motion training, these sparse but critical learning signals are further diluted by abundant regular motion samples, making it difficult for a unified controller to naturally acquire expert-level highly dynamic skills.

To improve highly dynamic motion performance, existing methods often adopt a specialist optimization paradigm and rely on preprocessed high-quality motion data~\cite{Liao2025BeyondMimic,Xie2025KungfuBot,Sleiman2026ZEST,Zhang2025Any2Track}. After smoothing, retargeting, and contact correction, such data usually provide stable and physically consistent reference trajectories, which are suitable for offline motion replay and specific skill reproduction. However, this paradigm is insufficient for unified humanoid control and online teleoperation. On the one hand, specialist optimization concentrates learning pressure on a narrow highly dynamic distribution, which can change the state-action mappings that support generalist behaviors and degrade fundamental motor abilities. On the other hand, skills learned from high-quality offline data do not directly adapt to low-quality reference inputs in online teleoperation. Inertial motion capture signals are often affected by timing errors, root motion drift, and pose inconsistency. These errors can be rapidly amplified during aerial phases, contact transitions, and landing recovery, leading to tracking failure. Therefore, unified humanoid control needs to jointly address highly dynamic capability expansion, generalist ability retention, and robustness to online reference inputs.

To build a generalist humanoid controller capable of online tracking of highly dynamic human motions, we propose Extreme-RGMT, a progressive highly dynamic skill learning framework. The framework starts from stable generalist motor competence and progressively expands toward expert-level highly dynamic skills. It consists of two stages: Stage~I establishes a generalist motion tracking base policy over a diverse multi-source motion distribution, and Stage~II expands this base policy toward highly dynamic capabilities.

Stage~I aims to provide a stable base controller for subsequent highly dynamic skill learning. To this end, we build on the controller design in~\cite{Ma2026RAGMT}, which uses dynamics-guided reference motion encoding and an asymmetric actor--critic policy structure. Our architectural enhancements separately encode proprioceptive and action histories, apply individual feature normalization, and regularize the aggregated command representation with FSQ~\cite{Mentzer2023FSQ}. These modifications improve the stability of the control representation. We train a generalist motion-tracking base policy over a diverse multi-source motion distribution. This base policy provides broad motion coverage and stable fundamental motor competence. Based on its tracking performance, we stratify the motion data into mastered motions and challenging highly dynamic motions, which provides the basis for asymmetric training in Stage~II.

In Stage~II, we design PACE, a Progressive Acquisition and Consolidation for Expansion mechanism, to mitigate the conflict between highly dynamic skill learning and generalist ability retention. Learning highly dynamic skills requires stronger optimization pressure on difficult motion segments, while directly reinforcing these segments may degrade previously mastered abilities. PACE therefore divides training into two roles: the acquisition branch focuses on challenging highly dynamic motions and provides stronger learning signals for expert-level dynamic segments; the consolidation branch uses reference policy regularization to constrain policy drift on mastered motions, thereby preserving fundamental motor abilities during specialist expansion.

Furthermore, we introduce Segment-Aware Trajectory Advantage Resampling (STAR) to address the scarcity of effective experience in highly dynamic motion training. For highly dynamic motions from inertial motion capture data, training often suffers from sparse informative experience, limited effective samples, and unstable learning signals from many failed trajectories. STAR uses the difficulty prior obtained from adaptive sampling to prioritize high-advantage trajectory fragments, allowing effective experience in high-failure temporal regions to contribute more fully to policy updates and improving sample utilization for highly dynamic skill learning.

Experimental results show that Extreme-RGMT achieves state-of-the-art generalist whole-body motion-tracking performance, including substantially improved completion of challenging highly dynamic motions. To the best of our knowledge, Extreme-RGMT is the first generalist humanoid controller to enable online teleoperation tracking of highly dynamic motions from inertial motion-capture inputs. This result extends highly dynamic humanoid control from offline replay of preprocessed motions to real-time human-driven control and advances generalist whole-body motion-tracking controllers toward highly dynamic motor capabilities at the human-expert level.

The main contributions of this work are summarized as follows:
\begin{itemize}
    \item We propose Extreme-RGMT, a two-stage progressive learning framework for generalist humanoid motion tracking. The framework first learns a generalist base policy from a diverse multi-source motion distribution, then stratifies motions into mastered and challenging sets based on tracking statistics to establish the basis for highly dynamic skill expansion.
    \item We introduce PACE, a role-specific acquisition and consolidation mechanism. It assigns challenging and mastered motions to distinct training roles, supporting highly dynamic skill acquisition while constraining drift from established generalist control behaviors.
    \item We develop Segment-Aware Trajectory Advantage Resampling (STAR), which uses difficulty priors to identify and resample high-advantage trajectory fragments, improving the utilization of scarce effective experience in highly dynamic motion learning.
    \item We conduct comprehensive simulation and hardware evaluations under fixed references and online inertial motion-capture inputs. Extreme-RGMT achieves state-of-the-art generalist whole-body motion-tracking performance and reliably executes challenging motions from both reference modalities.
\end{itemize}
%===============================================================================

\section{Related Work}

\subsection{Specialist Training for Challenging Humanoid Skills}
Specialist controllers have achieved reliable execution on a range of challenging humanoid tasks. Prior work has addressed agile locomotion over complex terrains and sparse footholds~\cite{He2025Attention,Rudin2025Parkour,Ma2026VPIES,Ma2026TerAdapt,Wang2025MoRE,WangH-RSS-25,Wang2026APEX}, as well as fall recovery from diverse postures~\cite{Huang2025StandingUp,He2025GettingUp}.
Specialist whole-body imitation has also enabled highly dynamic behaviors such as martial arts, backflips, and contact-intensive motions. KungfuBot~\cite{Xie2025KungfuBot} learns highly dynamic skills through physics-based whole-body control. BeyondMimic~\cite{Liao2025BeyondMimic} extends motion tracking toward versatile control through guided diffusion, while ZEST~\cite{Sleiman2026ZEST} investigates zero-shot embodied skill transfer for athletic robot control. OmniXtreme~\cite{Wang2026OmniXtreme} further targets highly dynamic humanoid control. These methods typically employ task-specific training environments, reward formulations, curricula, or reference-trajectory processing to attain reliable execution within their target settings. For highly dynamic motion imitation in particular, smoothing, retargeting, and contact correction can provide physically consistent reference trajectories that support offline replay and specialist skill reproduction.

However, the optimization objectives of such methods are closely tailored to particular motion categories, terrain environments, or isolated skill types. Their capability boundaries therefore remain largely determined by the target training distribution, making it difficult to jointly cover broad motion distributions.

\begin{figure*}[htbp]
    \centering
    \includegraphics[width=\textwidth]{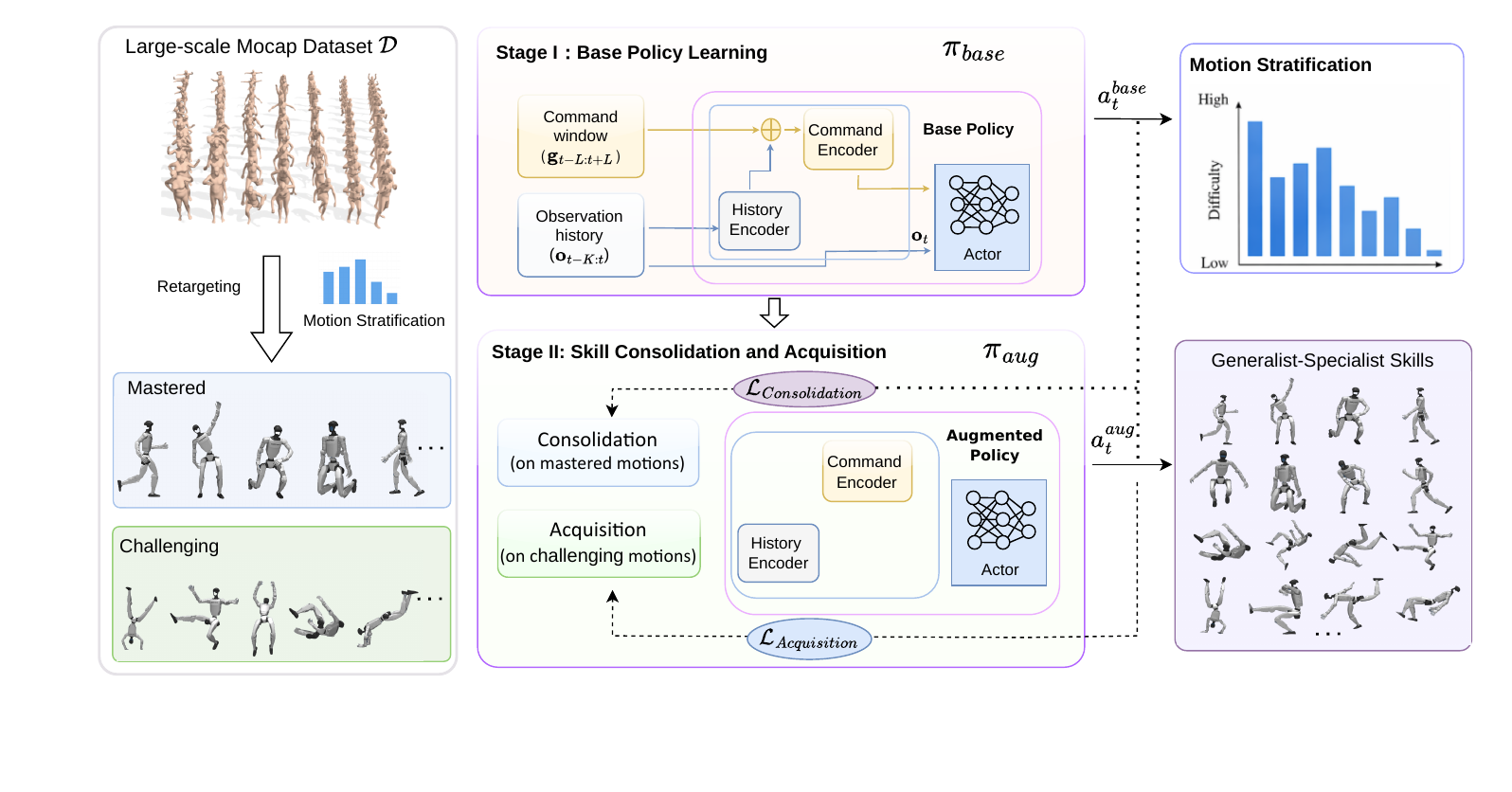}
    \vspace{-2cm}
    \caption{Two-stage progressive training framework. In Stage I, diverse multi-source motion-capture sequences are retargeted to the humanoid to train a generalist base policy, \(\pi_{\mathrm{base}}\), over the full motion distribution. The tracking performance of \(\pi_{\mathrm{base}}\) is then used to stratify motions into mastered and challenging sets. In Stage II, \(\pi_{\mathrm{base}}\) is augmented into \(\pi_{\mathrm{aug}}\) and optimized asymmetrically: mastered motions consolidate learned skills, while challenging motions drive the acquisition of specialized behaviors.}
    \label{fig:framework}
\end{figure*}

\subsection{Generalist Humanoid Motion Tracking}
Early learning-based whole-body motion tracking focused on stable reference-motion imitation for physical characters. VMP~\cite{Serifi2024VMP} proposed transferable motion priors to improve tracking robustness under disturbances. Expressive~\cite{Cheng2024Expressive} explored expressive whole-body control for humanoid robots, HumanPlus~\cite{Fu2024HumanPlus} demonstrated human-to-humanoid shadowing and imitation, and iCub3~\cite{Dafarra2024iCub3} introduced a humanoid avatar system for remote immersive embodiment. These works established a foundation for closed-loop whole-body control from human motion references.
Subsequent work has extended humanoid controllers in several directions. EGM~\cite{EGM_2025} improves training efficiency for highly dynamic whole-body control, while KungfuBot2~\cite{KungfuBot2_2025} and AMS~\cite{Pan2025AMS} explore joint learning of diverse whole-body skills and the use of heterogeneous data. Other work studies promptable behavior modeling with BFM-Zero~\cite{Li2025BFMZero}, scalable data collection with TWIST2~\cite{TWIST2_2025}, long-horizon closed-loop teleoperation with CLONE~\cite{Li2025CLONE}, exoskeleton-mediated interaction with HOMIE~\cite{Ben2025HOMIE}, and visually conditioned control with visual imitation~\cite{Allshire2025VisualImitation}, advancing humanoid control from offline motion reproduction toward more interactive and deployable settings.

As large-scale motion data~\cite{Mahmood2019AMASS,Harvey2020RobustMotion}, improved retargeting methods~\cite{Araujo2025GMR,Yang2025OmniRetarget}, and scalable training frameworks have developed, research has increasingly focused on using a single policy to cover broad motion distributions. GMT~\cite{GMT_2025} proposes a general whole-body motion-tracking framework, and UniTracker~\cite{UniTracker_2025} further learns a unified whole-body motion tracker. SONIC~\cite{Luo2025SONIC} advances more natural humanoid whole-body control by scaling motion-tracking training. RGMT~\cite{Ma2026RAGMT} combines the current dynamical state with a local reference window to improve tracking robustness under complex motion conditions.

Despite the continued expansion of motion coverage, training data scale, and conditioning modalities, rare highly dynamic behaviors, such as backflips, parkour transitions, and rapid contact switches, remain underrepresented in broad training distributions. Mixing such motions directly into the full distribution risks diluting their learning signal, whereas further optimization on challenging motions alone may degrade stable tracking over the original motion repertoire. Therefore, enabling a unified policy to reliably extend toward highly dynamic and contact-switching motions while retaining broad motion coverage remains a key challenge in generalist humanoid motion tracking.

\subsection{Progressive Skill Acquisition and Retention}
Progressive skill learning is closely related to continual reinforcement learning, which seeks to acquire new tasks while retaining previously learned capabilities. CLEAR~\cite{Rolnick2019CLEAR} reconciles stability and plasticity without explicit task boundaries by using off-policy experience replay to stabilize prior knowledge and on-policy learning to acquire new skills. To suppress forgetting, Global Alignment~\cite{Bai2024GlobalAlignment} preserves prior knowledge through global representation alignment. SAFE~\cite{Zhao2024SAFE} employs slow and fast parameter-efficient adaptation, while Bayesian continual learning~\cite{Lee2024Bayesian} organizes updates according to Bayesian principles.

Within reinforcement learning, CPPO~\cite{Zhang2024CPPO} studies continual policy optimization for learning new tasks. Studies of plasticity loss~\cite{Juliani2024Plasticity} and Churn~\cite{Tang2025Churn} analyze and mitigate reduced plasticity in on-policy reinforcement learning. Self-Composing Policies~\cite{Malagon2024SelfComposing} support scalable continual reinforcement learning through policy composition. Online World Models~\cite{Liu2025OnlineWorldModels} and Knowledge Retention~\cite{Fu2025KnowledgeRetention} further address continual learning through model-based planning and knowledge preservation. Related studies also investigate lifelong robotic reinforcement learning and continual learning in vision-language-action models~\cite{Meng2025RoboticLifelong,Zhang2026AtomicVLA,Hu2026SimpleRecipe,Liu2026VLAForgetting}. These studies provide important foundations for balancing the retention of prior knowledge with the acquisition of new capabilities over task sequences.

Highly dynamic humanoid skill expansion differs from general continual-learning settings because it occurs within a single whole-body tracking task over an imbalanced motion distribution. Challenging motions are scarce and failure prone, and their critical control requirements are concentrated in short highly dynamic segments. We therefore adopt PACE, an asymmetric acquisition and consolidation mechanism that organizes skill acquisition on challenging motions while using reference-policy regularization and a progress-adaptive consolidation weight to constrain policy drift on mastered motions. This design supports progressive expansion from broad generalist motions to highly dynamic behaviors.

\section{System Overview}
\label{sec:system_overview}

\subsection{Formulation}
\label{sec:formulation}

As shown in Fig.~\ref{fig:framework}, we use reinforcement learning to train a generalist humanoid motion-tracking controller 
that tracks reference human motions while progressively extending its capabilities to highly dynamic behaviors. 
We formulate this problem as a partially observable Markov decision process (POMDP) 
$\mathcal{M}=(\mathcal{S},\mathcal{O},\mathcal{A},P,r,\gamma)$. 
At each control step \(t\), the policy \(\pi_{\theta}\) receives a history of proprioceptive observations \(o^{\mathrm{prop}}_{t-H:t}\), the corresponding past actions \(a_{t-H-1:t-1}\), and a local reference-motion window \(g_{t-L:t+L}\). It outputs a residual joint-position command \(a_t\).

We optimize the policy with PPO~\cite{Schulman2017PPO} in parallel simulation~\cite{Makoviychuk2021IsaacGym} to maximize the expected tracking return. 
We first introduce Stage~I (detailed in Sec.~\ref{sec:generalist_tracking}), in which a generalist base policy $\pi_{\mathrm{base}}$ is learned to achieve broad motion coverage and robust fundamental motor skills. We then present Stage~II (detailed in Sec.~\ref{sec:progressive_expansion}), which augments $\pi_{\mathrm{base}}$ into $\pi_{\mathrm{aug}}$ to improve tracking of highly dynamic motions while retaining generalist capability.

\subsection{Control Interface}
\label{sec:control_interface}

At each policy step, the proprioceptive observation is defined as
\begin{equation}
o_t^{\mathrm{prop}} =
[\mathbf{g}_t^{\mathrm{proj}},\ \omega_t,\ q_t-q_0,\ \dot{q}_t],
\end{equation}
where $\mathbf{g}_t^{\mathrm{proj}}$ is the gravity direction projected into the base frame, $\omega_t$ is the
base angular velocity, $q_t-q_0$ is the joint-position offset from the nominal pose, and $\dot{q}_t$ is the joint
velocity. The actor uses $o_{t-H:t}^{\mathrm{prop}}$, the corresponding previous-action history $a_{t-H-1:t-1}$, and a reference-motion window
$g_{t-L:t+L}=[g_{t-L},\ldots,g_{t+L}]$, where
\begin{equation}
g_\tau =
[v_\tau^{\mathrm{ref}},\ \omega_\tau^{\mathrm{ref}},\
\mathbf{g}_\tau^{\mathrm{ref}},\ q_\tau^{\mathrm{ref}}].
\end{equation}
Here, $v_\tau^{\mathrm{ref}}$, $\omega_\tau^{\mathrm{ref}}$, $\mathbf{g}_\tau^{\mathrm{ref}}$, and
$q_\tau^{\mathrm{ref}}$ denote the reference base linear velocity, base angular velocity, gravity direction, and
joint pose, respectively. In our experiments, the actor uses a 10-frame proprioceptive and action history, and a
21-token reference window.

\begin{figure}[t]
\centering
\includegraphics[width=\columnwidth]{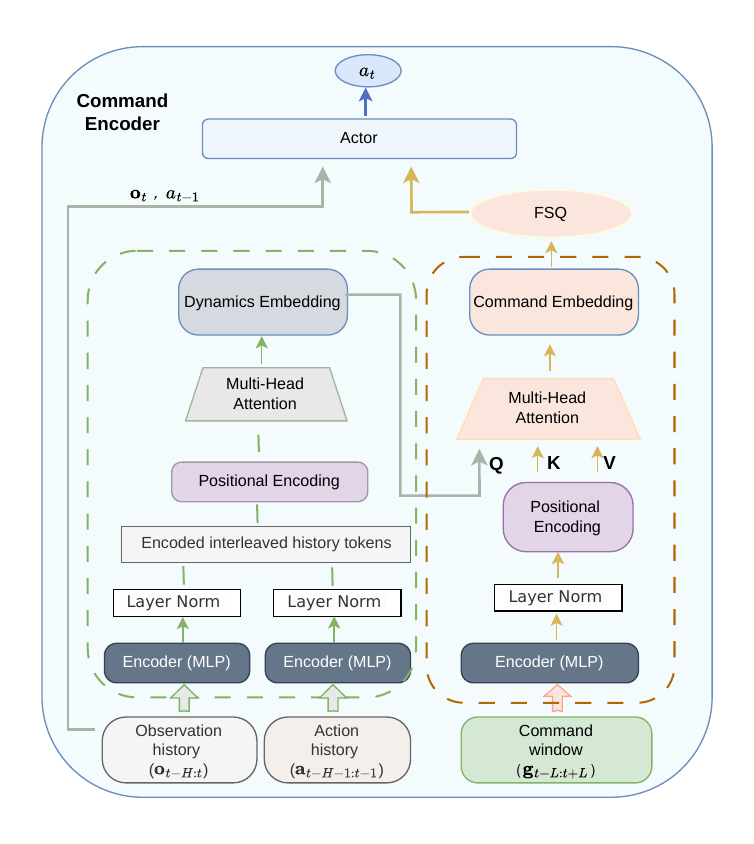}
\caption{Policy architecture for command encoding. Encoded proprioceptive and action histories are interleaved and processed by a causal history encoder to form the query for cross attention. The local reference motion window supplies the keys and values. FSQ regularizes the aggregated command feature before fusion in the actor.}
\label{fig:command_encoder}
\end{figure}

We use an asymmetric actor--critic setup. The actor only uses deployable observations, while the critic
additionally receives privileged robot and reference-state information, including the reference base height,
tracked body-link positions and orientations, and the robot base linear velocity.

The action space is a 29-dimensional residual joint-position command, $a_t\in\mathbb{R}^{29}$. The residual is
added to the reference joint pose to obtain the target joint position,
\begin{equation}
q_t^{\mathrm{tar}} = q_t^{\mathrm{ref}} + a_t .
\end{equation}
The low-level PD controller then generates joint torques as
\begin{equation}
\tau_t = K_p(q_t^{\mathrm{tar}} - q_t) - K_d \dot{q}_t .
\end{equation}

\section{Generalist Motion Tracking Training}
\label{sec:generalist_tracking}

We train a generalist base policy $\pi_{\mathrm{base}}$ on the full multi-source motion distribution in Stage~I. Its
goal is to establish broad motion-tracking coverage and robust fundamental control, providing the foundation for the
progressive skill expansion in Stage~II.

\subsection{Policy Architecture}
\label{sec:policy_architecture}

Building on the dynamics-guided command encoder~\cite{Ma2026RAGMT}, our policy architecture integrates recent
proprioceptive states, action history, and a local reference-motion command. Since these inputs convey distinct
physical information and operate at different temporal scales, they are processed by separate branches before fusion,
as illustrated in Fig.~\ref{fig:command_encoder}.

Following the control interface in Sec.~\ref{sec:control_interface}, the proprioceptive history
$o_{t-H:t}^{\mathrm{prop}}$ and the corresponding previous-action history $a_{t-H-1:t-1}$ are encoded by separate input branches. For each
history index $\tau\in[t-H,t]$, $o_{\tau}^{\mathrm{prop}}$ denotes the proprioceptive observation at time $\tau$,
and $a_{\tau-1}$ denotes the action executed before that observation. The two inputs are projected as
\begin{equation}
z_{\tau}^{o}=\mathrm{LN}_o(f_o(o_{\tau}^{\mathrm{prop}})),\qquad
z_{\tau-1}^{a}=\mathrm{LN}_a(f_a(a_{\tau-1})),
\end{equation}
where $\mathrm{LN}$ denotes LayerNorm~\cite{Ba2016LayerNorm}, and $f_o$ and $f_a$ are multi-layer perceptron
(MLP) input encoders for proprioceptive observations and actions, respectively. The encoded state and action
tokens are then arranged as an interleaved history sequence,
\begin{equation}
\mathcal{H}_t=[z_{t-H-1}^{a},z_{t-H}^{o},\ldots,z_{t-1}^{a},z_t^{o}],
\end{equation}
and passed to a causal history encoder, which only uses information available up to the current control step,
\begin{equation}
h_t=\mathrm{Enc}_{\mathrm{hist}}(\mathcal{H}_t).
\end{equation}
The action history provides recent closed-loop control context, while the separate state and action encoders keep
their feature scales stable before temporal aggregation.

The local reference-motion window is encoded by an independent command branch. For each reference token,
\begin{equation}
e_\tau^g=\mathrm{LN}_g(f_g(g_\tau))+p_\tau,
\end{equation}
where $f_g$ is an MLP input encoder for reference-command tokens, $p_\tau$ is the temporal positional embedding, and
$Z_t^g=[e_{t-L}^g,\ldots,e_{t+L}^g]$ denotes the encoded command-token sequence. We apply LayerNorm independently
to the proprioceptive, action, and reference-command branches, rather than empirical normalization based on running
observation statistics commonly used in legged locomotion policies~\cite{Zhou2025ALARM,Rudin2022Learning}. By normalizing
each branch in feature space, this design accommodates both the diverse motion distribution in generalist whole-body
tracking and the distribution shifts that arise during highly dynamic motion learning.

The history representation serves as the query in a dynamics-guided cross-attention module that aggregates the
encoded command tokens,
\begin{equation}
u_t=\mathrm{CrossAttn}(Q=W_qh_t,\ K=Z_t^g,\ V=Z_t^g).
\end{equation}
Here, $Q$, $K$, and $V$ denote the query, key, and value features, respectively. This mechanism conditions reference
aggregation on the robot's recent proprioceptive and control history, enabling the policy to attend to the portions of
the local reference window that are most relevant to its current control state. Such state-dependent selection is
particularly important for highly dynamic motions, where small phase deviations can substantially alter the relevant
reference information.

To regularize the history-conditioned command representation, we introduce a finite scalar quantization (FSQ)
bottleneck~\cite{Mentzer2023FSQ}:
\begin{equation}
\hat{u}_t=\mathcal{Q}_{\mathrm{FSQ}}(u_t).
\end{equation}
Specifically, $u_t$ is factorized into two 32-dimensional tokens and quantized using FSQ. Applied after
history-conditioned command aggregation, rather than directly to the raw reference inputs, the FSQ bottleneck
constrains the information delivered to the actor to a discrete and bounded latent representation. This reduces
sensitivity to local inconsistencies in highly dynamic reference trajectories and provides a structured command
representation for subsequent skill expansion.

Finally, the actor receives the current proprioceptive observation $o_t^{\mathrm{prop}}$, the previous action
$a_{t-1}$, and the quantized command representation $\hat{u}_t$, and outputs a residual joint-position command:
\begin{equation}
a_t=\pi_\theta(o_t^{\mathrm{prop}},a_{t-1},\hat{u}_t).
\end{equation}
Network dimensions and optimization settings are summarized in Table~\ref{tab:training_hyperparameters}.
The critic follows the asymmetric setup described in Sec.~\ref{sec:control_interface} and estimates the value from
privileged training information.

\subsection{Training Setup}

\subsubsection{Reward Design}
Both Stage~I and the acquisition environments of Stage~II optimize motion tracking with PPO under a shared reward
formulation. Following~\cite{Liao2025BeyondMimic,Ma2026RAGMT}, the reward combines motion-imitation terms with
physical regularizers, as summarized in Table~\ref{tab:reward_terms}. The imitation terms encourage matching of the
reference body pose and velocity, whereas the regularizers penalize unstable or physically implausible behaviors.

\begin{table}[t]
\centering
\footnotesize
\caption{Reward Components Used for Motion Tracking}
\label{tab:reward_terms}
\begin{tabular}{llr}
\toprule
Type & Term & Weight \\
\midrule
Tracking & global anchor orientation & 0.5 \\
Tracking & relative body position & 1.0 \\
Tracking & relative body orientation & 1.0 \\
Tracking & global body linear velocity & 1.0 \\
Tracking & global body angular velocity & 1.0 \\
Regularization & action rate & -0.1 \\
Regularization & joint position limits & -10.0 \\
Regularization & undesired contacts & -0.1 \\
Regularization & feet slip & -0.1 \\
\bottomrule
\end{tabular}
\end{table}

\subsubsection{Domain Randomization}
To improve robustness and sim-to-real transfer, we apply the same simulation perturbation protocol during Stage~I
and in the acquisition environments of Stage~II. This protocol includes dynamics randomization, proprioceptive
observation noise, and reference-command perturbations. The corresponding ranges are listed in
Table~\ref{tab:domain_randomization}.

\begin{table}[t]
\centering
\scriptsize
\setlength{\tabcolsep}{3pt}
\caption{Domain-Randomization Ranges Used During Training}
\label{tab:domain_randomization}
\begin{tabular}{p{0.22\columnwidth}p{0.34\columnwidth}p{0.30\columnwidth}}
\toprule
Category & Quantity & Range \\
\midrule
Dynamics & ground friction & $[0.10,1.75]$ \\
Dynamics & added base mass & $[-3,6]$ kg \\
Dynamics & base CoM offset & $x\in[-0.025,0.025]$ m; $y,z\in[-0.05,0.05]$ m \\
Dynamics & motor strength scale & $[0.8,1.2]$ \\
Dynamics & PD gain scale & $[0.8,1.2]$ \\
Dynamics & motor zero offset & $[-0.01,0.01]$ rad \\
Dynamics & joint armature scale & $[1.0,1.05]$ \\
Dynamics & external push interval & $[1,3]$ s \\
\midrule
Observation noise & gravity & 0.05 \\
Observation noise & angular velocity & 0.2 rad/s \\
Observation noise & joint position & 0.01 rad \\
Observation noise & joint velocity & 0.5 rad/s \\
\midrule
Command perturbation & base linear velocity & $\pm0.5$ m/s \\
Command perturbation & base angular velocity & $\pm0.52$ rad/s \\
Command perturbation & gravity direction & 0.05 \\
Command perturbation & joint pose & $\pm0.1$ rad \\
\bottomrule
\end{tabular}
\end{table}

\subsubsection{Adaptive Motion Sampling}
\label{sec:adaptive_motion_sampling}
We employ adaptive motion sampling~\cite{Liao2025BeyondMimic,Ma2026RAGMT} to allocate more rollout
initializations to temporal bins that are difficult to track. Each motion sequence is partitioned into temporal bins,
and bins with higher tracking errors or failure frequencies are sampled more frequently. Each bin maintains an
exponential moving average of failure or high-error events:
\begin{equation}
c_i \leftarrow (1-\alpha)c_i+\alpha f_i ,
\end{equation}
where $f_i$ denotes the observed failure statistic for bin $i$. The bin scores are clipped, normalized, and mixed
with a uniform baseline:
\begin{equation}
s_i=\mathrm{Normalize}(\mathrm{clip}(c_i,0,c_{\max})),\qquad
\hat{p}_i \propto s_i + \epsilon_u/N ,
\end{equation}
where $N$ is the number of bins and $\epsilon_u$ denotes the uniform-baseline ratio. Adaptive sampling is applied to
the full motion set in Stage~I and to the challenging motion set $\mathcal{D}_c$ in the acquisition environments of
Stage~II. Meanwhile, the consolidation environments of Stage~II sample uniformly from the mastered motion set
$\mathcal{D}_m$ to maintain broad coverage of previously acquired behaviors.

\begin{table}[t]
\centering
\footnotesize
\setlength{\tabcolsep}{3pt}
\renewcommand{\arraystretch}{1.05}
\caption{Architecture and Optimization Hyperparameters}
\label{tab:training_hyperparameters}
\begin{tabular}{>{\raggedright\arraybackslash}p{0.48\columnwidth}
                >{\raggedright\arraybackslash}p{0.44\columnwidth}}
\toprule
\multicolumn{2}{l}{\textit{Policy architecture}} \\
% Latent dimension & 64 \\
State encoder dimensions & $[64,128,64]$ \\
Action encoder dimensions & $[29,64,64]$ \\
Command encoder dimensions & $[38,128,64]$ \\
Actor hidden dimensions & $[1024,1024,512,256]$ \\
Critic hidden dimensions & $[1024,1024,512,512]$ \\
\midrule
\multicolumn{2}{l}{\textit{PPO optimization}} \\
% Hyperparameter & Value \\
Rollout horizon & 24 steps per environment \\
PPO epochs per update & 5 \\
Mini-batches per update & 4 \\
Initial learning rate & $1\times10^{-3}$, adaptive KL schedule \\
Target KL divergence & 0.01 \\
Discount factor $\gamma$ & 0.99 \\
GAE parameter $\lambda_{\mathrm{GAE}}$ & 0.95 \\
PPO clipping parameter & 0.2 \\
Entropy coefficient & 0.005 \\
Acquisition fraction $\xi$ & 0.8 \\
\bottomrule
\end{tabular}
\end{table}

\subsection{Motion Stratification}
\label{sec:motion_stratification}

The full motion set $\mathcal{D}$ consists of retargeted motions from LAFAN1~\cite{Harvey2020RobustMotion}
and AMASS~\cite{Mahmood2019AMASS}, together with in-house inertial motion-capture recordings collected using an
Xsens system~\cite{XsensMVNAnimate}. All motion sequences are retargeted to the Unitree G1 morphology and
resampled to 50 Hz to match the policy control frequency. Table~\ref{tab:dataset_sources} reports the duration
of each source dataset used for generalist motion tracking.

\begin{table}[t]
\centering
\footnotesize
\caption{Source-Level Motion Dataset Durations}
\label{tab:dataset_sources}
\begin{tabular}{lrrr}
\toprule
Source & Min & Hours & Share \\
\midrule
LAFAN1 & 146.651 & 2.444 & 78.94\% \\
AMASS & 30.677 & 0.511 & 16.51\% \\
In-house & 8.446 & 0.141 & 4.55\% \\
\midrule
Total & 185.774 & 3.096 & 100.00\% \\
\bottomrule
\end{tabular}
\end{table}

After learning $\pi_{\mathrm{base}}$, we partition each motion sequence longer than 10~s into 10-s clips while
retaining shorter sequences as individual clips. Each resulting clip is evaluated using five randomized rollouts,
with its completion rate defined as the fraction of successful rollouts, and clips attaining a completion rate of at
least 80\% are assigned to the mastered motion set $\mathcal{D}_m$, whereas the remaining clips are assigned to the
challenging motion set $\mathcal{D}_c$. These disjoint sets jointly cover all motion clips and subsequently support
capability consolidation and highly dynamic skill acquisition, respectively.

\begin{figure}[t]
\centering
\includegraphics[width=0.85\columnwidth]{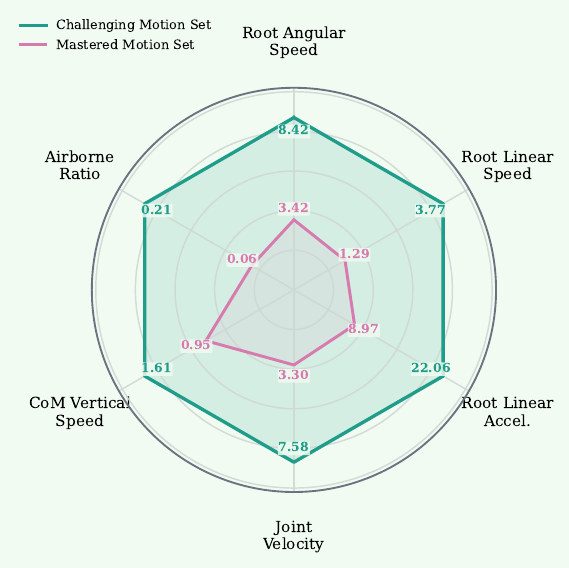}
\caption{Kinematic distribution comparison between the mastered and challenging motion sets. The kinematic axes
report the median of the per-motion 99th-percentile statistics for both motion sets.}
\label{fig:motion_distribution}
\end{figure}

The stratification yields a mastered motion set $\mathcal{D}_m$ and a challenging motion set $\mathcal{D}_c$,
where $\mathcal{D}_m$ contains most of the motion data and supports capability consolidation, whereas the relatively
smaller $\mathcal{D}_c$ is used for focused skill acquisition. Their durations and training roles are reported in
Table~\ref{tab:motion_sets}, while Fig.~\ref{fig:motion_distribution} further characterizes their differences by
comparing their kinematic distributions. The mastered set $\mathcal{D}_m$ primarily comprises regular motions with
lower root and joint dynamic measures and a lower airborne ratio, representing the broad motion regime that the base
policy can track reliably. In contrast, $\mathcal{D}_c$ exhibits stronger root and joint dynamics together with more
frequent airborne motion, indicating control regimes that require dedicated skill acquisition. Accordingly,
$\mathcal{D}_m$ and $\mathcal{D}_c$ serve as the consolidation and acquisition sets, respectively, during progressive
expansion.

\begin{table}[t]
\centering
\footnotesize
\caption{Role-Specific Motion Sets Used During Training}
\label{tab:motion_sets}
\begin{tabular}{llrp{0.32\columnwidth}}
\toprule
Motion set & Symbol & Hours & Role \\
\midrule
Mastered & $\mathcal{D}_m$ & 2.82 & Consolidation and broad coverage \\
Challenging & $\mathcal{D}_c$ & 0.28 & Highly dynamic acquisition \\
\bottomrule
\end{tabular}
\end{table}

\section{Progressive Highly Dynamic Skill Expansion}
\label{sec:progressive_expansion}

Building on the base policy $\pi_{\mathrm{base}}$ and the stratified motion sets introduced in
Sec.~\ref{sec:motion_stratification}, we train an augmented policy $\pi_{\mathrm{aug}}$ in Stage~II. This stage aims
to acquire highly dynamic skills while retaining the broad motion-tracking capability established in Stage~I, using
PACE for asymmetric acquisition and consolidation and STAR to prioritize high-advantage trajectory fragments.

\subsection{PACE: Progressive Acquisition and Consolidation for Expansion}
\label{sec:pace}

In Stage~II, PACE progressively augments the generalist motion tracker into a unified controller that retains broad
motion coverage while developing specialist competence in highly dynamic motion regimes. It addresses the
stability-plasticity trade-off by coupling specialist acquisition on challenging motions with consolidation over
mastered motions. Specifically, PACE allocates parallel environments asymmetrically between these two roles and
adjusts the consolidation strength according to training progress. Algorithm~\ref{alg:stage2} summarizes the overall
PACE procedure, and the following sections describe its key components.

\subsubsection{Environment Allocation for Consolidation and Acquisition}
To retain established generalist capability during highly dynamic skill acquisition, Stage~II asymmetrically
allocates parallel simulation environments across two training roles. A fraction $\xi$ of the environments is assigned
to skill acquisition, while the remaining fraction $1-\xi$ is used for capability consolidation. During training,
consolidation environments sample uniformly from the mastered motion set $\mathcal{D}_m$ to maintain coverage of
stabilized behaviors, whereas acquisition environments sample adaptively from the challenging motion set
$\mathcal{D}_c$ to focus training resources on motions that continue to limit policy performance. Since highly dynamic
motions often terminate early and thereby reduce the number of valid training samples, we set $\xi=0.8$ to ensure
sufficient effective experience for skill acquisition.

\begin{algorithm}[t]
\caption{PACE: Progressive Acquisition and Consolidation for Expansion}
\label{alg:stage2}
\begin{algorithmic}[1]
\Require $\pi_{\mathrm{base}}$, $\mathcal{D}_m$, $\mathcal{D}_c$, $\xi$, $\beta$, $\lambda_{\mathrm{base}}$, $\kappa$, $\rho_{\mathrm{ref}}$
\State $\pi_{\mathrm{ref}} \gets \pi_{\mathrm{base}}$, $\pi_\theta \gets \pi_{\mathrm{base}}$, $\bar{\rho} \gets \rho_{\mathrm{ref}}$
\For{$k=0,1,\ldots$}
    \State Split environments into $\mathcal{E}_A$ ($\xi$) and $\mathcal{E}_C$ ($1-\xi$)
    \State Sample $\mathcal{D}_m$ uniformly in $\mathcal{E}_C$
    \State Sample $\mathcal{D}_c$ adaptively in $\mathcal{E}_A$
    \State Evaluate $\mathcal{L}_{\mathrm{con}}$ on $\mathcal{E}_C$ against $\pi_{\mathrm{ref}}$
    \State Update $\bar{\rho}$ and $\lambda_{\mathrm{con}}$ from $\rho$
    \State Evaluate $\mathcal{L}_{\mathrm{acq}}$ on $\mathcal{E}_A$
    \State Update $\theta$ using $\mathcal{L}_{\mathrm{acq}}+\lambda_{\mathrm{con}}\mathcal{L}_{\mathrm{con}}$
\EndFor
\State $\pi_{\mathrm{aug}} \gets \pi_\theta$
\State \Return $\pi_{\mathrm{aug}}$
\end{algorithmic}
\end{algorithm}

\subsubsection{Asymmetric Objective Formulation}
\label{sec:asymmetric_objective}
In Stage~II, the trainable policy $\pi_\theta$ is initialized with the parameters of the base policy
$\pi_{\mathrm{base}}$ learned in Stage~I and optimized across two environment groups with distinct roles. The challenging
motion set $\mathcal{D}_c$ provides reinforcement-learning signals for specialist acquisition, whereas the mastered
motion set $\mathcal{D}_m$ provides consolidation signals that constrain policy drift. Accordingly, all loss terms are
written in minimization form:
\begin{equation}
\min_{\theta}\ 
\left\{
\mathcal{L}_{\mathrm{Acquisition}}^{\mathcal{D}_c}
+ \lambda_{\mathrm{con}}^{t}\mathcal{L}_{\mathrm{Consolidation}}^{\mathcal{D}_m}
\right\}.
\end{equation}
Rollouts from acquisition environments contribute only to $\mathcal{L}_{\mathrm{Acquisition}}^{\mathcal{D}_c}$,
whereas samples from consolidation environments contribute only to
$\mathcal{L}_{\mathrm{Consolidation}}^{\mathcal{D}_m}$. Let $d_{\mathcal{D}_m}$ denote the state distribution induced
by consolidation samples collected from motions in $\mathcal{D}_m$. Within the consolidation environments of
Stage~II, the base policy $\pi_{\mathrm{base}}$ learned in Stage~I serves as the reference policy
$\pi_{\mathrm{ref}}$. This policy is not updated during Stage~II and provides an alignment target for the control
outputs of the trainable policy on mastered motions. For a state $s$, let $a_\theta(s)$ and $a_{\mathrm{ref}}(s)$
denote the actions output by $\pi_\theta$ and $\pi_{\mathrm{ref}}$, respectively. The consolidation objective is
defined as
\begin{equation}
\mathcal{L}_{\mathrm{Consolidation}}^{\mathcal{D}_m} = \mathbb{E}_{s\sim d_{\mathcal{D}_m}}
\left[\left\|a_\theta(s)-a_{\mathrm{ref}}(s)\right\|_2^2\right],
\end{equation}
which constrains the discrepancy between the current and reference policy actions to preserve stable control over
mastered motions. On the challenging motion set,
$\mathcal{L}_{\mathrm{Acquisition}}^{\mathcal{D}_c}$ is optimized using the clipped PPO policy loss~\cite{Schulman2017PPO}:
\begin{equation}
\mathcal{L}_{\mathrm{PPO}}
=
-\mathbb{E}_{t}
\left[
\min\left(
r_tA_t,\tilde{r}_tA_t
\right)
\right],
\end{equation}
where $\tilde{r}_t=\mathrm{clip}(r_t,1-\epsilon,1+\epsilon)$ is the clipped policy ratio, and $A_t$ is the advantage
estimate. Together, these objectives update a single augmented policy that acquires highly dynamic skills while
retaining established generalist capability.

\subsubsection{Progress-Adaptive Consolidation Weighting}
Highly dynamic motions frequently terminate early during the initial stage of training, leaving the acquisition
environments with few valid samples that provide useful learning signals. If the consolidation constraint is assigned
excessive weight at this stage, the policy is overly constrained near the base policy and cannot make the control
adjustments needed to acquire highly dynamic skills. As training progresses, the acquisition environments produce more
valid samples, indicating that challenging motions are providing more stable optimization signals. Increasing the
consolidation strength at this point supports continued specialist acquisition while limiting interference with
mastered motions.

We therefore use the realized ratio of valid acquisition samples as an indicator of training progress and adapt the
consolidation weight $\lambda_{\mathrm{con}}^t$ accordingly. At update $t$, let $N_A^t$ and $N_C^t$ denote the
numbers of valid samples collected from acquisition and consolidation environments, respectively, and define the
acquisition-sample ratio as
\begin{equation}
\rho^{(t)} = \frac{N_A^{t}}{N_A^{t}+N_C^{t}}.
\end{equation}
To reduce fluctuations in sample counts across individual updates, we maintain an exponential moving average
initialized at $\rho_{\mathrm{ref}}$:
\begin{equation}
\bar{\rho}^{(t)} = \beta\bar{\rho}^{(t-1)} + (1-\beta)\rho^{(t)},
\qquad \bar{\rho}^{(0)}=\rho_{\mathrm{ref}}.
\end{equation}
The consolidation weight is then computed as
\begin{equation}
\lambda_{\mathrm{con}}^{t} = \min\left(1.0,\;
\lambda_{\mathrm{base}} + \kappa\max\left(0,\bar{\rho}^{(t)}-\rho_{\mathrm{ref}}\right)\right).
\end{equation}
When the smoothed acquisition-sample ratio does not exceed $\rho_{\mathrm{ref}}$, the consolidation weight remains
at its base value $\lambda_{\mathrm{base}}$, leaving sufficient optimization capacity for highly dynamic skill
acquisition. Once the ratio exceeds this threshold, the consolidation strength increases with gain $\kappa$ and is
capped at $1.0$. We use $\lambda_{\mathrm{base}}=0.3$, $\kappa=5.0$, $\rho_{\mathrm{ref}}=0.6$, and $\beta=0.99$.

\subsection{STAR: Segment-Aware Trajectory Advantage Resampling}
\label{sec:star}

In the acquisition environments of Stage~II, training focuses on the challenging motion set $\mathcal{D}_c$, but the
effective control demands of highly dynamic motions are often concentrated in a small number of critical temporal
regions. This issue is particularly pronounced for challenging highly dynamic motions captured with the Xsens system,
where lower reference-data quality increases the frequency of early termination and further reduces the availability of
effective trajectory experience. Although adaptive motion sampling increases the visitation frequency of temporal bins
with high failure statistics, it allocates sampling opportunities only according to bin-level difficulty and cannot
distinguish trajectory attempts of differing quality within the same difficult temporal region. Consequently, merely
increasing the sampling frequency of high-failure bins can still leave PPO updates influenced by a large number of
low-value failed trajectories.

To make better use of the scarce effective experience in difficult motions, we introduce Segment-Aware Trajectory
Advantage Resampling (STAR). STAR first transforms bin-level difficulty information from adaptive sampling into a
transition-level difficulty prior and uses it for difficulty-conditioned advantage normalization. It then evaluates the
learning potential of contiguous trajectory fragments using raw advantages, selects high-advantage fragments, and
resamples their transitions into PPO mini-batches with higher probability. This mechanism enables acquisition-side
updates to focus on trajectory experience with greater improvement potential in high-difficulty temporal regions,
thereby improving sample utilization for highly dynamic skill learning.

\subsubsection{Difficulty-Conditioned Advantage Normalization}
Adaptive motion sampling produces a bin-level distribution $p=[p_1,\ldots,p_B]$ over the temporal bins of the
challenging motion set $\mathcal{D}_c$, where $B$ denotes the number of bins. To propagate this bin-level difficulty
information to subsequent advantage normalization and trajectory-fragment selection, STAR defines a transition-level
difficulty weight. For a rollout transition $t$, let $b_t$ denote its associated reference bin. Its difficulty weight
is defined by scaling the bin probability relative to the uniform baseline $1/B$:
\begin{equation}
w_t=Bp_{b_t}.
\end{equation}
When $w_t>1$, the reference bin associated with transition $t$ has a sampling probability above the uniform baseline,
indicating higher failure frequency or optimization demand under the current policy. This weight is subsequently used
for difficulty-conditioned advantage normalization and high-advantage trajectory-fragment resampling.

Given the transition-level difficulty weights, STAR computes the raw advantage for each transition in the acquisition
rollout buffer $D$ as
\begin{equation}
A_t^{\mathrm{raw}}=R_t-V_t.
\end{equation}
Here, $R_t$ is the value target obtained using generalized advantage estimation (GAE)~\cite{Schulman2016GAE}, and
$V_t$ is the critic's value estimate at transition $t$. STAR then separates transitions whose difficulty weights
exceed the uniform baseline from the remaining transitions:
\begin{equation}
H=\{t\in D\mid w_t>1\}, \qquad
E=\{t\in D\mid w_t\le 1\}.
\end{equation}
The advantage distributions of these two groups can differ in both mean and scale. A shared normalization statistic
can therefore obscure the relative learning potential of transitions in high-difficulty regions. STAR instead
normalizes raw advantages independently within $H$ and $E$:
\begin{equation}
A_t =
\begin{cases}
\dfrac{A_t^{\mathrm{raw}}-\mu_H}{\sigma_H+\epsilon}, & t\in H, \\[6pt]
\dfrac{A_t^{\mathrm{raw}}-\mu_E}{\sigma_E+\epsilon}, & t\in E.
\end{cases}
\end{equation}
Here, $\mu_H,\sigma_H$ and $\mu_E,\sigma_E$ are the means and standard deviations of raw advantages in the
high-difficulty and remaining groups, respectively. The normalized advantage $A_t$ is used for PPO updates, so that
the advantage scale in high-difficulty regions is determined by its own distribution. The raw advantage
$A_t^{\mathrm{raw}}$ is retained to evaluate the improvement potential of contiguous trajectory fragments.

\subsubsection{Trajectory-Fragment Advantage Scoring}
After obtaining the high-difficulty group $H$, STAR identifies high-advantage trajectory fragments associated with
high-difficulty bins and collects their transitions into a sampling pool for mixed mini-batch construction. A trajectory
fragment is defined as the maximal contiguous transition sequence from a single environment within the current rollout,
delimited by termination boundaries. Let $\tau(t)$ denote the fragment containing transition $t$.

To ensure coverage across difficult temporal regions, STAR evaluates candidate trajectory fragments separately within
each high-difficulty bin. For a high-difficulty bin $b$ and a fragment $\tau$, the associated high-difficulty
transitions are defined as
\begin{equation}
S_{b,\tau}
=
\{t\in H\mid b_t=b,\ \tau(t)=\tau\}.
\end{equation}
When $S_{b,\tau}\neq\varnothing$, STAR uses the average raw advantage of its transitions to measure the improvement
potential of fragment $\tau$ within bin $b$:
\begin{equation}
q_{b,\tau}
=
\frac{1}{|S_{b,\tau}|}
\sum_{t\in S_{b,\tau}} A_t^{\mathrm{raw}}.
\end{equation}
A larger $q_{b,\tau}$ indicates that the contiguous fragment contains transitions with greater improvement potential
in the corresponding difficult bin.

For each high-difficulty bin $b$, STAR independently ranks valid bin-fragment pairs by $q_{b,\tau}$ and retains the
top
\begin{equation}
k_b=
\max\left(\left\lceil \rho_{\mathrm{topk}} n_b \right\rceil,1\right)
\end{equation}
pairs, where $n_b$ is the number of valid pairs for bin $b$ and $\rho_{\mathrm{topk}}=0.05$ is the retention ratio.
This bin-wise selection ensures that each difficult temporal region can retain fragments with learning value without
being dominated by bins containing more samples. The selected fragment set $T^\star$ is formed by the union of
fragments associated with the retained pairs across all high-difficulty bins. STAR then constructs a sampling pool from
the current acquisition rollout buffer:
\begin{equation}
P=\{t\in D\mid \tau(t)\in T^\star\}.
\end{equation}

\subsubsection{Difficulty-Weighted Mixed Mini-Batches}
Given the selected-fragment sampling pool $P$, STAR forms each acquisition PPO mini-batch by combining transitions
from $P$ with standard samples from the acquisition rollout buffer $D$. Let $M$ denote the mini-batch size and let
$\rho_{\mathrm{star}}=0.25$ denote the resampling ratio. The number of transitions sampled from $P$ is
\begin{equation}
M_{\mathrm{star}}
=
\min\left(M,\max\left(\lfloor \rho_{\mathrm{star}}M \rfloor,1\right)\right).
\end{equation}
If $P$ is empty, no resampling is performed.

STAR assigns each selected trajectory fragment $\tau$ an average difficulty score:
\begin{equation}
\eta_\tau =
\frac{1}{|\{t\in D\mid \tau(t)=\tau\}|}
\sum_{t\in D:\tau(t)=\tau} w_t.
\end{equation}
For a transition $t\in P$, the sampling weight is determined by the score of its associated fragment:
\begin{equation}
\omega_t=
\frac{\eta_{\tau(t)}}{\sum_{j\in P}\eta_{\tau(j)}},
\qquad t\in P.
\end{equation}
STAR samples $M_{\mathrm{star}}$ transitions from $P$ according to $\omega_t$ and draws the remaining
$M-M_{\mathrm{star}}$ transitions from $D$. The resulting mini-batch is used to update the acquisition-side PPO
objective.

%===============================================================================

% \input{ppo_sampling_method}

% \section{Implementation and Training Details}
% \label{sec:implementation}
\begin{table*}[t]
    \caption{Generalist and specialist motion-tracking results. Tracking-error metrics are reported as mean $\pm$ standard deviation over five random seeds. $\mathrm{E}_{\mathrm{MPJPE}}$, $d_{\mathrm{vel}}$, and $d_{\mathrm{acc}}$ are reported in mm, mm/frame, and mm/frame$^2$, respectively.}
    \label{tab:motion_tracking_capability}
    \centering
    \scriptsize
    \setlength{\tabcolsep}{1.5pt}
    \newcommand{\stdplaceholder}[1]{\textcolor{gray}{#1}^{\dagger}}
    \begin{tabular*}{\textwidth}{@{\extracolsep{\fill}}lcccccccc}
        \toprule
        \rowcolor{generalistblue}
        \multicolumn{9}{c}{{(a) Generalist Capability}} \\
        \midrule
        Method & \multicolumn{4}{c}{In-source Motion} & \multicolumn{4}{c}{Unseen Motion} \\
        \cmidrule(lr){2-5} \cmidrule(lr){6-9}
        & Succ. (\%) $\uparrow$ & $\mathrm{E}_{\mathrm{MPJPE}}$ $\downarrow$ & $d_{\mathrm{vel}}$ $\downarrow$ & $d_{\mathrm{acc}}$ $\downarrow$
        & Succ. (\%) $\uparrow$ & $\mathrm{E}_{\mathrm{MPJPE}}$ $\downarrow$ & $d_{\mathrm{vel}}$ $\downarrow$ & $d_{\mathrm{acc}}$ $\downarrow$ \\
        \midrule
        ExBody2~\cite{Ji2025ExBody2} & 85.63 & $69.08 \pm 1.07$ & $5.10 \pm {0.20}$ & $2.75 \pm {0.11}$ & 66.78 & $103.54 \pm 2.27$ & $7.25 \pm {0.32}$ & $3.68 \pm {0.18}$ \\
        BeyondMimic~\cite{Liao2025BeyondMimic} & 94.72 & $43.41 \pm 0.53$ & $4.35 \pm {0.13}$ & $2.35 \pm {0.08}$ & 71.34 & $69.17 \pm 1.36$ & $5.25 \pm {0.24}$ & $2.78 \pm {0.13}$ \\
        SONIC~\cite{Luo2025SONIC} & 99.33 & $40.09 \pm 0.14$ & $3.89 \pm {0.08}$ & $\textbf{2.03} \pm {0.03}$ & 93.67 & $47.39 \pm 0.33$ & $4.22 \pm {0.17}$ & $2.24 \pm {0.09}$ \\
        RGMT~\cite{Ma2026RAGMT} & 99.12 & $40.14 \pm {0.18}$ & $3.75 \pm {0.09}$ & $2.12 \pm {0.06}$ & 94.58 & $46.56 \pm 0.26$ & $4.08 \pm {0.13}$ & $2.20 \pm {0.07}$ \\
        \rowcolor{stageoneblue}
        Extreme-RGMT (Stage I) & 99.54 & $\textbf{40.07} \pm {0.13}$ & $\textbf{3.73} \pm {0.07}$ & $2.11 \pm {0.04}$ & 95.13 & $\textbf{45.80} \pm {0.22}$ & $\textbf{4.00} \pm {0.11}$ & $\textbf{2.17} \pm {0.06}$ \\
        \rowcolor{fullgreen}
        Extreme-RGMT (Full) & \textbf{99.76} & $40.79 \pm 0.21$ & $3.76 \pm {0.08}$ & $2.14 \pm {0.05}$ & \textbf{96.68} & $46.91 \pm 0.28$ & $4.03 \pm {0.12}$ & $2.20 \pm {0.08}$ \\
        \midrule
        \rowcolor{specialistgreen}
        \multicolumn{9}{c}{{(b) Specialist Capability}} \\
        \midrule
        Method & \multicolumn{4}{c}{XtremeMotion} & \multicolumn{4}{c}{AMASS Challenging Motions} \\
        \cmidrule(lr){2-5} \cmidrule(lr){6-9}
        & Succ. (\%) $\uparrow$ & $\mathrm{E}_{\mathrm{MPJPE}}$ $\downarrow$ & $d_{\mathrm{vel}}$ $\downarrow$ & $d_{\mathrm{acc}}$ $\downarrow$
        & Succ. (\%) $\uparrow$ & $\mathrm{E}_{\mathrm{MPJPE}}$ $\downarrow$ & $d_{\mathrm{vel}}$ $\downarrow$ & $d_{\mathrm{acc}}$ $\downarrow$ \\
        \midrule
        OmniXtreme~\cite{Wang2026OmniXtreme} & \textbf{100.00} & $\textbf{38.71} \pm {0.14}$ & $\textbf{3.22} \pm {0.04}$ & $\textbf{2.07} \pm {0.03}$ & 36.16 & $68.26 \pm {2.18}$ & $5.99 \pm {0.31}$ & $3.43 \pm {0.18}$ \\
        Fine-Tuning & 71.43 & $44.37 \pm {0.88}$ & $4.16 \pm {0.15}$ & $2.69 \pm {0.09}$ & 54.55 & $53.72 \pm {1.72}$ & $5.63 \pm {0.28}$ & $2.84 \pm {0.15}$ \\
        \rowcolor{stageoneblue}
        Extreme-RGMT (Stage I) & 21.42 & $46.72 \pm {0.97}$ & $4.21 \pm {0.17}$ & $2.76 \pm {0.13}$ & 18.18 & $55.17 \pm {1.95}$ & $5.42 \pm {0.30}$ & $2.87 \pm {0.16}$ \\
        \rowcolor{fullgreen}
        Extreme-RGMT (Full) & \textbf{100.00} & $40.18 \pm {0.19}$ & $3.56 \pm {0.10}$ & $2.25 \pm {0.08}$ & \textbf{90.91} & $\textbf{46.39} \pm {0.83}$ & $\textbf{4.15} \pm {0.16}$ & $\textbf{2.32} \pm {0.07}$ \\
        \bottomrule
    \end{tabular*}
\end{table*}

\section{Experiments}
\label{sec:result}

This section evaluates Extreme-RGMT from four perspectives: generalist motion tracking, highly dynamic skill expansion, mechanism analysis, and real-world deployment. First, we compare its performance on generalist motion tracking against representative methods. Next, we evaluate its ability to track highly dynamic motions and analyze the relationship between generalist capability retention and specialist capability expansion. We then examine the effects of the individual components of the proposed method through ablation studies. Finally, we deploy the policy on the Unitree G1 and evaluate its real-world execution performance under fixed-reference replay and Xsens-based online teleoperation.

\subsection{Generalist and Specialist Motion Tracking}

\subsubsection{Evaluation Metrics}
All evaluations are conducted in MuJoCo~\cite{Todorov2012MuJoCo}, and all quantitative results are averaged over five random seeds unless otherwise specified. We assess tracking performance using complementary completion, pose, and motion-fidelity metrics. 
Each rollout is initialized from the first frame of the reference motion and continues until the reference terminates or the policy fails. 

Completion performance is measured by the success rate (Succ.), defined as the proportion of rollouts that complete the reference motion. A rollout is deemed unsuccessful when the robot root height deviates from the reference by more than $0.2$~m.
To quantify pose-tracking accuracy, we report the root-relative mean per-joint euclidean position error $\mathrm{E}_{\mathrm{MPJPE}}$ (mm). To assess physical fidelity, we further calculate the joint velocity error $d_{\mathrm{vel}}$ (mm/frame) and joint acceleration error $d_{\mathrm{acc}}$ (mm/frame$^2$), which measure discrepancies in the first- and second-order joint motion, respectively.

\subsubsection{Generalist Capability Comparison}
To evaluate the general motion capability established in Stage~I and its retention after highly dynamic skill expansion, 
we compare methods on two general motion test subsets. 
The In-source Motion subset comprises motions from LAFAN1 and AMASS, 
and measures broad tracking performance within the training data sources.
The Unseen Motion subset is reconstructed from independently collected videos 
and includes common behaviors such as walking, dance, and martial arts. 
This subset evaluates cross-source generalization under changes in motion style, data source, 
and video-reconstruction artifacts.

We benchmark against four advanced generalist motion-tracking controllers: ExBody2~\cite{Ji2025ExBody2}, 
BeyondMimic~\cite{Liao2025BeyondMimic}, SONIC~\cite{Luo2025SONIC}, and RGMT~\cite{Ma2026RAGMT}. 
The publicly released SONIC checkpoint is evaluated directly. The respective implementations of ExBody2, BeyondMimic, 
and RGMT are retrained on exactly the same Stage-I corpus as Extreme-RGMT, 
yielding controlled comparisons under a common training-data condition. 
Because the original BeyondMimic formulation does not support large-scale joint optimization over multiple motions, 
its implementation is extended for multi-motion training.
For our method, Extreme-RGMT (Stage I) denotes the generalist base policy $\pi_{\mathrm{base}}$ 
before highly dynamic skill expansion, whereas Extreme-RGMT (Full) denotes the final augmented policy $\pi_{\mathrm{aug}}$ 
after Stage-II highly dynamic skill expansion. 
Reporting both variants directly assesses whether highly dynamic skill learning degrades the generalist capability established in Stage~I.

Table~\ref{tab:motion_tracking_capability}~(a) characterizes the stability of each method under changes in motion source. 
ExBody2 and BeyondMimic exhibit larger degradation on Unseen Motion, 
indicating that their joint-level tracking performance is more dependent on the training motion distribution. 
SONIC, RGMT, and Extreme-RGMT show smaller cross-source changes. 
Compared with RGMT trained on the same data, Extreme-RGMT (Stage I) consistently 
improves on both in-source and unseen motions. 
This gain stems from the policy design in Sec.~\ref{sec:policy_architecture}: the policy preserves RGMT's dynamics-conditioned selection of relevant commands from a local reference window, while separately encoding proprioceptive history, action history, and reference commands. Feature-space normalization establishes a stable interface across these streams, and FSQ regularizes the aggregated command representation. Together, these components reduce the influence of broad motion distributions and local reference inconsistencies on the control input, supporting stable joint-level tracking on unseen motions.

From Extreme-RGMT (Stage I) to Extreme-RGMT (Full), 
Stage-II training on challenging highly dynamic motions further improves success rates on both in-source and unseen motions, 
with a larger gain on unseen motions. This result indicates improved handling of large state variations, 
accumulated local errors, and contact transitions, thereby increasing execution stability for cross-source motions. 
The final Extreme-RGMT policy incurs slight increases in joint-position, joint-velocity, and joint-acceleration errors relative to the Stage-I base policy, reflecting the limited joint-level tracking-fidelity cost of highly dynamic skill expansion. Overall, Stage~II improves motion completion and cross-source generalization while retaining the broad general joint-tracking capability acquired in Stage~I.

\subsubsection{Specialist Capability Comparison}
To evaluate specialist tracking of highly dynamic motions, we use two test sets: XtremeMotion and AMASS Challenging Motions. XtremeMotion is the highly dynamic motion dataset publicly released with OmniXtreme, whereas AMASS Challenging Motions consists of difficult motions selected from AMASS. Both sets contain the same types of highly dynamic behaviors, with representative examples such as Butterfly Kick, Webster Flip, and Aerial Cartwheel. However, XtremeMotion motions undergo higher-quality preprocessing, resulting in reference trajectories with greater physical consistency and executability than directly retargeted AMASS motions. Together, these test sets evaluate adaptation to changes in reference quality for the same types of highly dynamic motions.

We evaluate four complementary policies to isolate the effect of specialist expansion: Extreme-RGMT (Stage I), Fine-Tuning, OmniXtreme~\cite{Wang2026OmniXtreme}, and Extreme-RGMT (Full). Extreme-RGMT (Stage I) is the generalist base policy before exposure to specialist highly dynamic training and provides the pre-expansion reference. Fine-Tuning is initialized from the same base policy and directly optimized over the full set of highly dynamic motions without invoking any of the methods proposed for our Stage~II. OmniXtreme is evaluated using its publicly released checkpoint. Extreme-RGMT (Full) denotes the final policy obtained after the complete Stage-II training procedure.

Table~\ref{tab:motion_tracking_capability}~(b) shows that the Stage-I policy attains low success rates on both test sets, indicating the difficulty of acquiring highly dynamic skills through general motion training alone. As motion diversity and dynamic difficulty increase, joint optimization is susceptible to gradient interference and conservative averaging, making it difficult for a single policy to satisfy the local control requirements of different highly dynamic behaviors. Although Fine-Tuning improves motion completion through dedicated training on the challenging motions, its overall performance remains limited.

On the higher-quality XtremeMotion set, both OmniXtreme and Extreme-RGMT (Full) reliably complete the evaluated motion set, while OmniXtreme attains lower joint-level errors. This result reflects OmniXtreme's high-precision tracking capability on its target motion library. When evaluation shifts to directly retargeted AMASS Challenging Motions, the relative performance changes substantially. OmniXtreme undergoes a marked reduction in success rate, whereas Extreme-RGMT (Full) attains the strongest motion completion and tracking fidelity. The two test sets cover matched categories of highly dynamic motions while differing in reference generation and preprocessing quality. The resulting performance change indicates stronger robustness of Extreme-RGMT to variations in reference quality.

\subsubsection{Generalist--Specialist Trade-off}
\begin{figure}[t]
\centering
\includegraphics[width=\columnwidth]{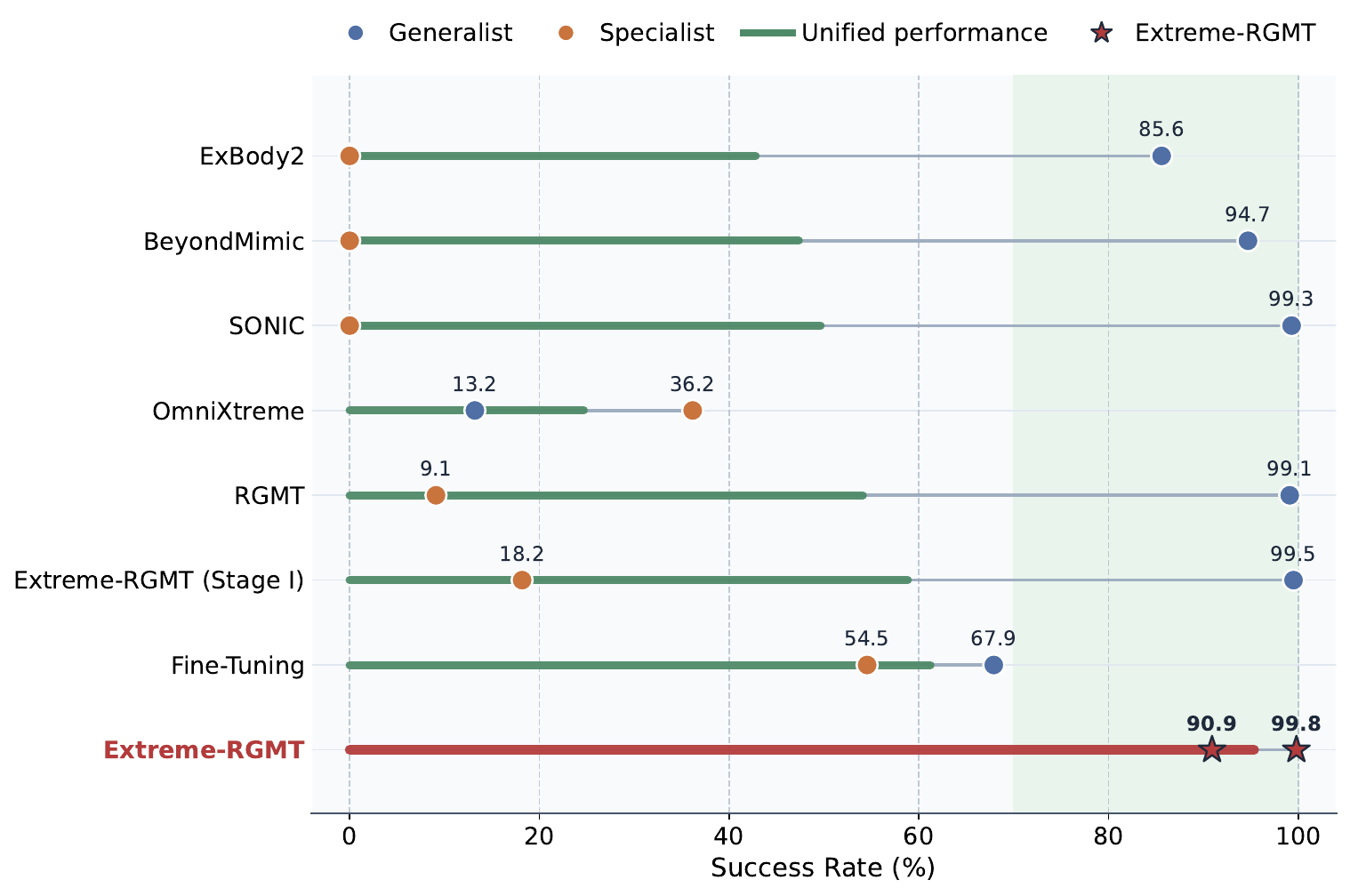}
\caption{Generalist--specialist capability trade-off. Blue and orange markers denote success rates on In-source Motion and AMASS Challenging Motions, respectively. The green bar denotes the equally weighted unified performance over the two evaluations.}
\label{fig:generalist_specialist_tradeoff}
\end{figure}

Fig.~\ref{fig:generalist_specialist_tradeoff} illustrates the performance relationship between generalist motion tracking and highly dynamic motion tracking across different methods. Blue and orange markers denote the generalist success rate on In-source Motion and the specialist success rate on AMASS Challenging Motions, respectively. The green bar reports the equally weighted average of the two success rates, providing a direct comparison of overall capability.

The methods exhibit different capability profiles. ExBody2, BeyondMimic, and SONIC achieve high success rates on general motions, but their performance on challenging highly dynamic motions remains limited. OmniXtreme shows a certain level of specialist tracking capability, whereas its generalist performance is relatively weak. Extreme-RGMT (Stage I) provides strong generalist motion-tracking capability but has not yet developed sufficient capability for highly dynamic motions. Direct Fine-Tuning improves specialist performance, but this improvement is accompanied by a reduction in generalist capability. After Stage-II training, Extreme-RGMT further improves the success rate on highly dynamic motions while maintaining a high generalist success rate, leading to the highest equally weighted average in the figure. These results indicate that the second-stage training of Extreme-RGMT expands the executable range of highly dynamic motions while preserving the established generalist motion-tracking capability.

\subsection{Ablation and Analysis}
\label{sec:ablation_study}

\subsubsection{Motion and Control Characteristics}
Fig.~\ref{fig:motion_distribution} and Table~\ref{tab:motion_sets} have shown that the challenging motion set $\mathcal{D}_c$ exhibits stronger dynamic characteristics than the mastered motion set $\mathcal{D}_m$, while being substantially smaller in scale. This distributional imbalance means that highly dynamic motions impose greater demands on instantaneous coordination, contact transitions, and landing recovery, yet their effective optimization signals can be diluted by the much larger distribution of regular motions.

\begin{figure}[t]
\centering
\includegraphics[width=0.98\columnwidth]{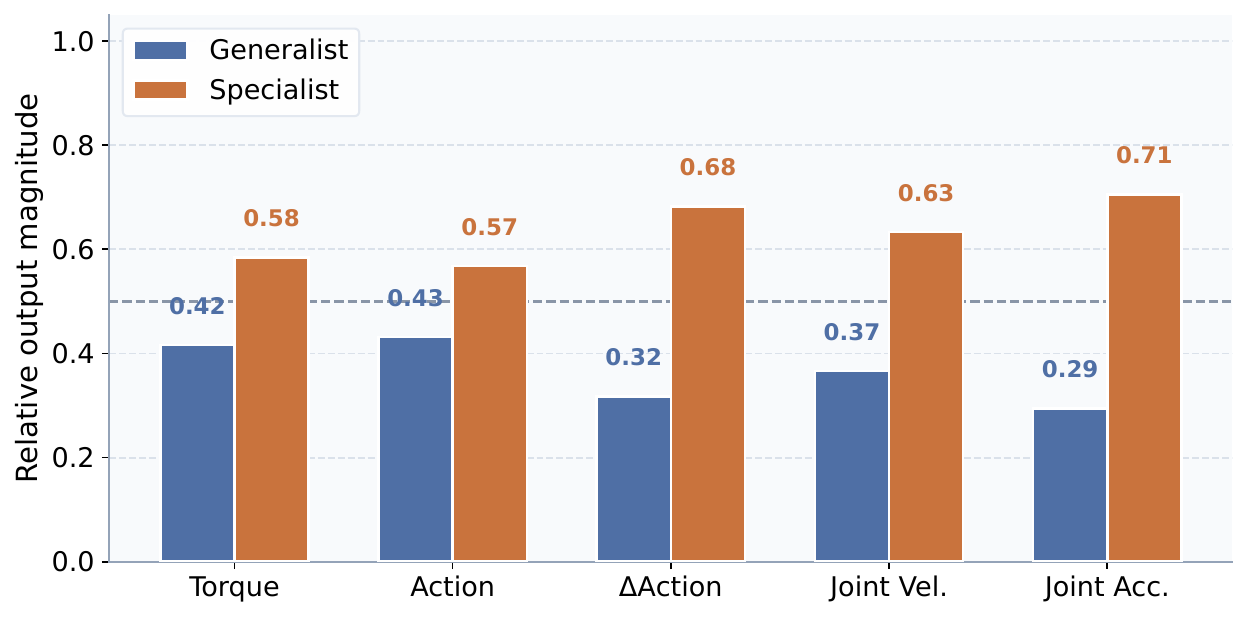}
\caption{Relative control-output magnitudes on Generalist and Specialist Motions. For each metric, the two values are normalized by their sum.}
\label{fig:output_difference}
\end{figure}

Fig.~\ref{fig:output_difference} further characterizes this difference at the control level. We compare the relative output magnitudes of the same augmented policy on Generalist and Specialist Motions, and observe that Specialist Motions induce larger action changes, joint velocities, and joint accelerations. This indicates that highly dynamic behaviors are not only more extreme in motion statistics, but also impose stronger instantaneous control demands on the policy output. Therefore, when directly optimizing over the full motion distribution, the large number of regular motion samples may dominate the optimization direction and bias the policy toward smoother and more conservative control patterns, making it difficult to satisfy the local control requirements of highly dynamic motions. Conversely, direct adaptation to the challenging set alone may disrupt the established generalist capability.

This imbalance imposes two requirements on a generalist controller that acquires highly dynamic skills. First, the policy should preserve the broad control patterns covered by $\mathcal{D}_m$. Second, it should obtain sufficient optimization signals from the rare but critical highly dynamic temporal regions in $\mathcal{D}_c$. To meet these requirements, PACE uses a consolidation constraint to limit policy drift on $\mathcal{D}_m$, while STAR prioritizes high-advantage trajectory fragments during training, improving the utilization of effective trajectory experience in $\mathcal{D}_c$.

\begin{figure*}[t]
\centering
\includegraphics[width=0.90\textwidth]{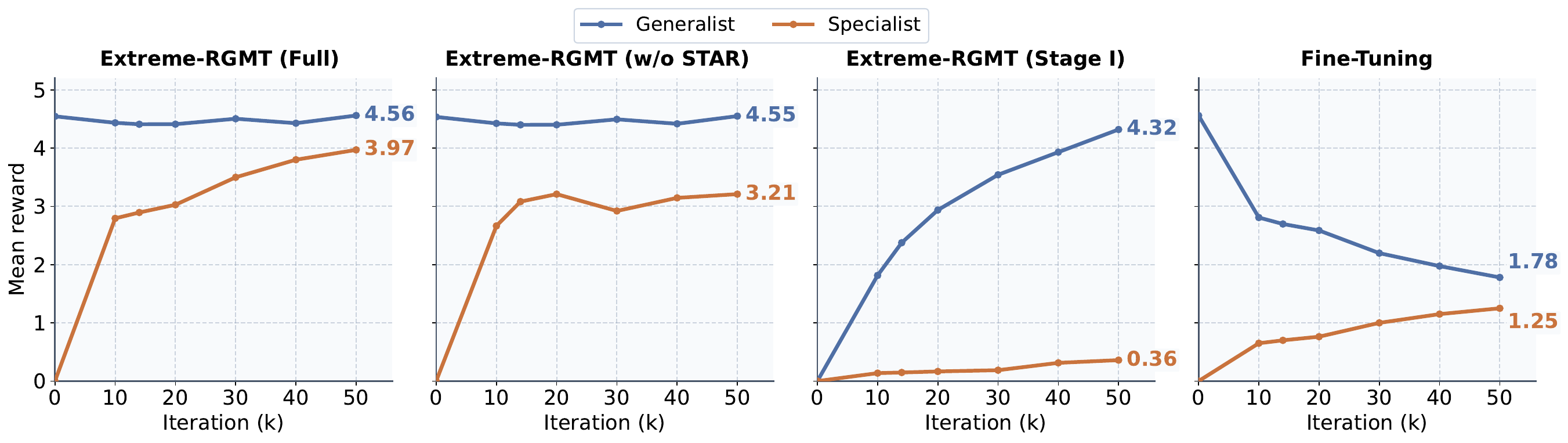}
\caption{Acquisition and retention dynamics during Stage-II highly dynamic skill expansion.}
\label{fig:progressive_acquisition}
\end{figure*}

\begin{figure*}[t]
\centering
\includegraphics[width=0.85\textwidth]{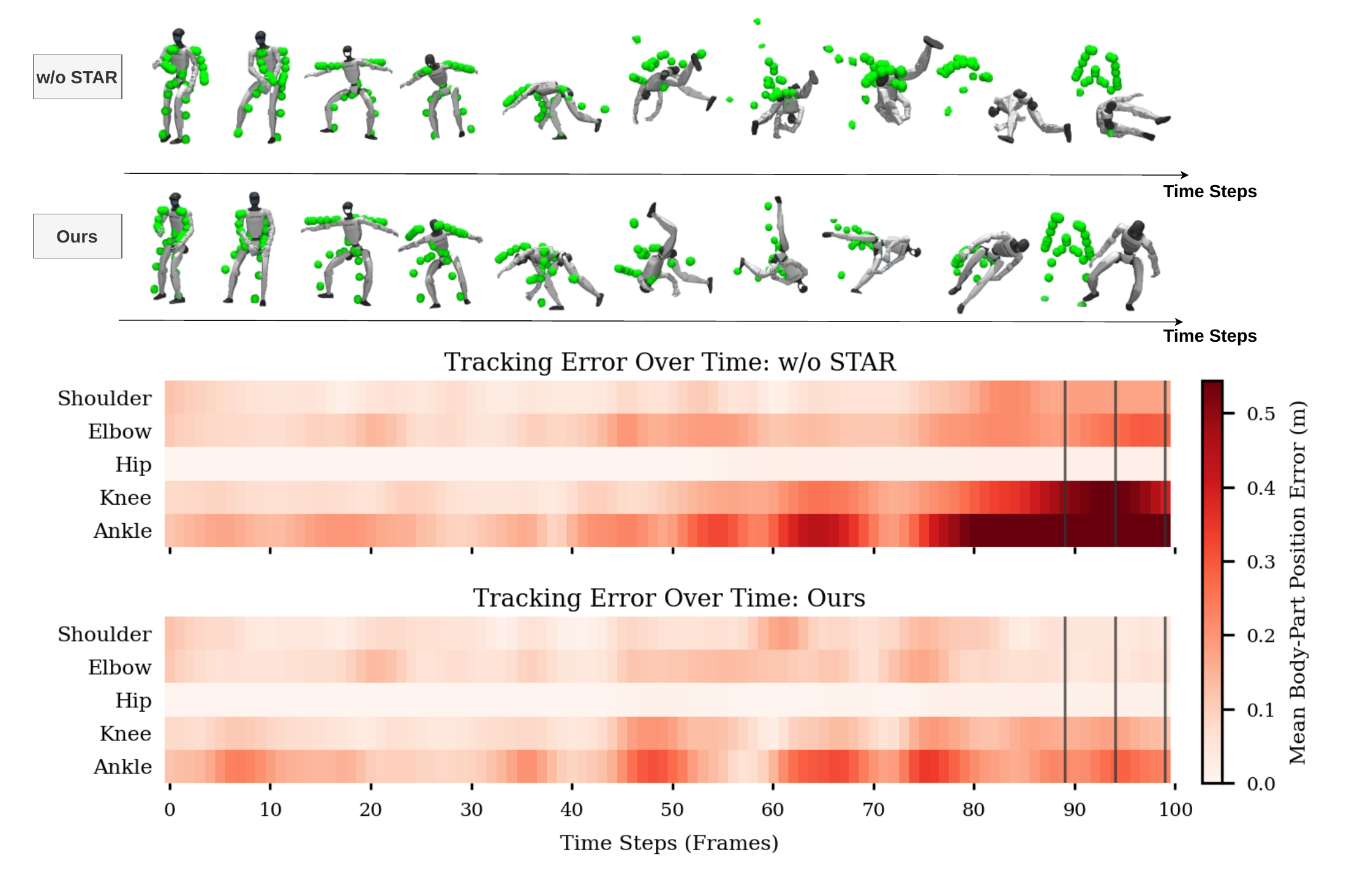}
\caption{Temporal tracking-error comparison on an aerial body twist from Xsens. STAR reduces error peaks in failure-critical frames, where aerial adjustment and landing recovery require concentrated control.}
\label{fig:star_temporal_heatmap}
\end{figure*}

% \begin{figure*}[!t]
% \centering
% \begin{minipage}{\textwidth}
% \centering
% \includegraphics[width=0.85\textwidth]{60_heatmap.pdf}\\[-0.5ex]
% (a) Aerial body twist
% \end{minipage}
% \vspace{0.1ex}
% \begin{minipage}{\textwidth}
% \centering
% \includegraphics[width=0.85\textwidth]{data/62_heatmap.pdf}\\[-0.5ex]
% (b) Backflip
% \end{minipage}
% \caption{Temporal tracking-error comparison on representative challenging motions. STAR reduces error peaks in failure-critical frames, where aerial adjustment and landing recovery require concentrated control.}
% \label{fig:star_temporal_heatmap}
% \end{figure*}

\subsubsection{Acquisition and Retention Training Dynamics}
Fig.~\ref{fig:progressive_acquisition} shows the evolution of Generalist and Specialist performance over training for different methods. Extreme-RGMT (Stage I) gradually improves Generalist performance, whereas its Specialist performance remains close to the initial level. This result shows that the broad control patterns learned in Stage I can cover the regular motion distribution, but the coordination, contact transitions, and recovery required for highly dynamic motions do not emerge from general motion training alone. As differences in dynamic difficulty and local control requirements increase, joint optimization becomes susceptible to gradient interference and conservative averaging, making it difficult to acquire sufficient specialist control capability for highly dynamic motions.

Fine-Tuning starts from the same generalist base policy and directly continues optimization on the highly dynamic data. Although it improves Specialist performance, its Generalist performance decreases markedly over training. This behavior suggests that direct specialist adaptation alters the established state--action mapping, progressively biasing the policy toward the challenging motion distribution and thereby weakening its ability to stably track the original broad motion repertoire.

In contrast, the complete Extreme-RGMT maintains stable Generalist capability throughout training while continuously improving Specialist capability and ultimately attaining the highest Specialist result. This outcome suggests that PACE limits excessive deviation from the base policy during highly dynamic skill acquisition, such that learning new skills does not come at the expense of established generalist capability.

Finally, the comparison with Extreme-RGMT (w/o STAR) further clarifies the role of STAR. Under the same acquisition and consolidation mechanism, removing STAR still allows the policy to retain high Generalist capability and improve Specialist capability during the early stage of training, although its Specialist performance subsequently plateaus. Compared with the continued improvement of the complete method in later training, this result indicates that sampling guided solely by difficulty at the bin-level does not fully exploit the critical learning signals in difficult motions. By prioritizing trajectory fragments with high advantages, STAR strengthens learning in temporal regions critical to failures, thereby improving the policy's ability to acquire difficult highly dynamic behaviors.

\subsubsection{PACE Ablation}
\begin{figure}[t]
\centering
\includegraphics[width=\columnwidth]{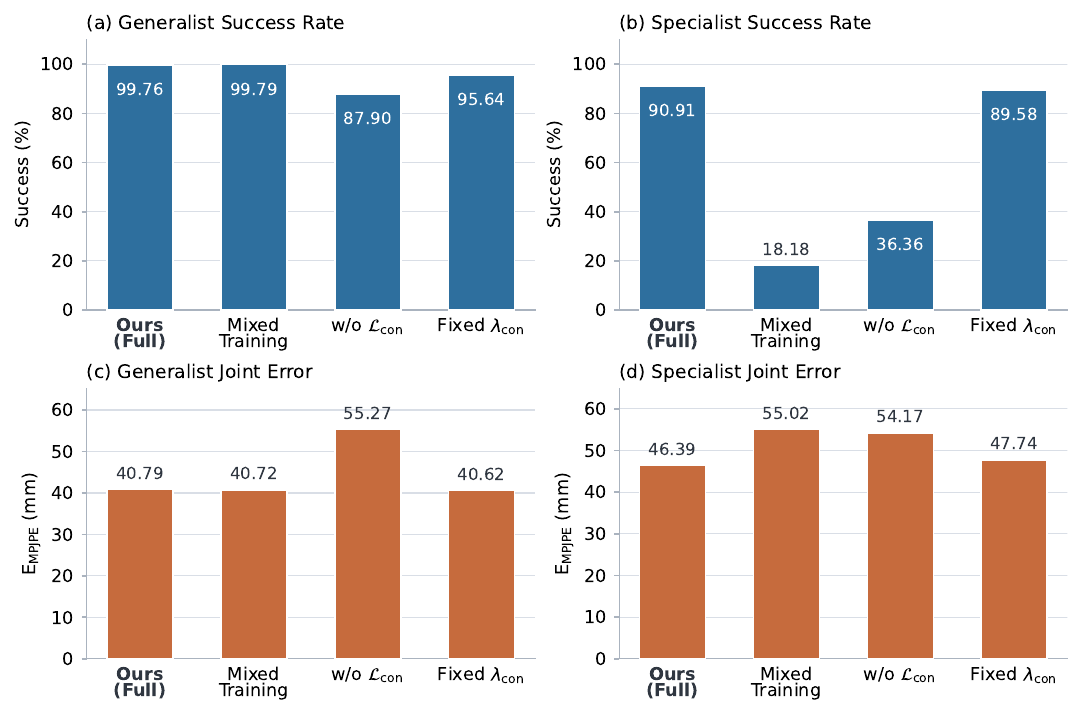}
\caption{PACE ablation results. (a) and (b) report Generalist and Specialist success rates, respectively. (c) and (d) report the corresponding joint-position errors $\mathrm{E}_{\mathrm{MPJPE}}$.}
\label{fig:pace_ablation_metrics}
\end{figure}

Fig.~\ref{fig:pace_ablation_metrics} analyzes the roles of training organization, the consolidation constraint, and its weighting strategy in PACE. Under the same Stage-II training budget, Mixed Training combines the mastered and challenging motion sets and continues training only with the PPO objective. The w/o $\mathcal{L}_{\mathrm{con}}$ variant retains the acquisition and consolidation environment split but removes the consolidation constraint with respect to the base policy. The Fixed $\lambda_{\mathrm{con}}$ variant retains the complete mechanism, while replacing the progress-adaptive consolidation weight with a fixed value of $\lambda_{\mathrm{con}}=0.5$.

Although Mixed Training retains favorable Generalist performance, it provides limited improvement on Specialist motions. This result indicates that, when optimization continues over the original mixed motion distribution, regular motion samples remain the dominant training signal and provide insufficient guidance for challenging highly dynamic motions. Removing $\mathcal{L}_{\mathrm{con}}$ decreases success rates and increases joint tracking errors in both evaluations. These degradations indicate that environment allocation alone is insufficient to preserve established capability reliably, whereas constraining the current policy to remain aligned with the base policy on mastered motions limits policy drift during specialist adaptation and mitigates interference with generalist motion tracking. The Fixed $\lambda_{\mathrm{con}}$ variant retains most of the performance, whereas the complete method achieves higher success rates in both evaluations and lower Specialist joint-position error.

\subsubsection{STAR Analysis}
\begin{table}[!t]
\caption{Specialist motion-tracking success rates (\%) by motion source.}
\label{tab:star_source_results}
\centering
\renewcommand{\arraystretch}{1.18}
\begin{tabular*}{\columnwidth}{@{\extracolsep{\fill}}lccc}
\toprule
Source & w/o STAR & Ours (Full) & Gain \\
\midrule
AMASS Motions~\cite{Mahmood2019AMASS} & 82.2 & 90.9 & $+8.7$ \\
In-house Xsens Motions~\cite{XsensMVNAnimate} & 45.5 & 86.3 & $+40.8$ \\
\bottomrule
\end{tabular*}
\end{table}

The earlier comparison in Fig.~\ref{fig:progressive_acquisition} between the complete method and Extreme-RGMT (w/o STAR) shows that, with acquisition and consolidation held fixed, STAR primarily affects the continued improvement of Specialist capability in later training, while the Generalist retention trends remain broadly similar. To further analyze this difference, Table~\ref{tab:star_source_results} compares Specialist success rates with and without STAR across motion sources. STAR improves performance on both AMASS and Xsens motions, with a substantially larger gain on Xsens motions. Without STAR, the success rate on Xsens motions is considerably lower than that on AMASS motions, whereas STAR substantially narrows this gap. Compared with AMASS, which is based on optical motion-capture data, Xsens inertial capture data have substantially lower quality and are more susceptible to root drift, local pose inconsistencies, and contact-timing errors. Consequently, their highly dynamic segments are more prone to tracking failures and rely more heavily on the effective use of limited informative experience.

Fig.~\ref{fig:star_temporal_heatmap} further examines the effect of STAR on critical temporal regions using an aerial body twist from Xsens. The complete method exhibits smaller tracking-error peaks during aerial body adjustment and landing recovery, which require rapid whole-body coordination and contact transitions and where local reference deviations can accumulate into tracking failures. By prioritizing trajectory fragments with high advantages that are associated with temporal bins with high failure rates, STAR strengthens effective learning signals near these critical regions, thereby reducing local error accumulation and improving the completion of difficult highly dynamic motions.

\begin{figure*}[t]
    \centering
    \includegraphics[width=0.88\textwidth,keepaspectratio]{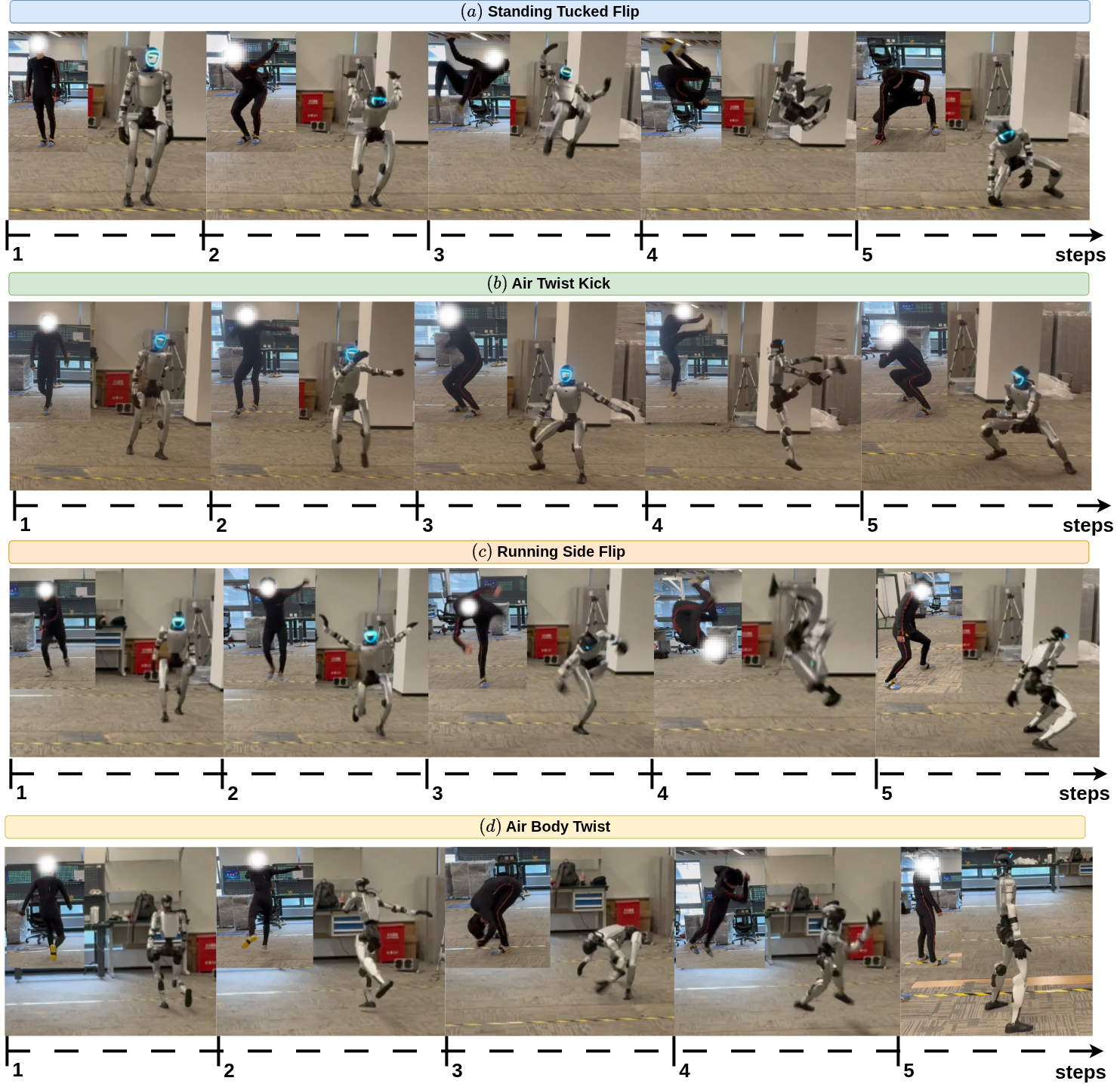}
    \caption{Real-world online teleoperation results for highly dynamic motions on the Unitree G1 humanoid platform.}
    \label{fig:real_world_high_dynamic}
\end{figure*}

\subsection{Real-World Evaluation and Deployment}

We deploy the complete policy on the Unitree G1 humanoid to evaluate execution under fixed highly dynamic references and online inertial motion-capture inputs. The robot has 29 actuated degrees of freedom. The policy operates at 50 Hz, and the low-level PD controller operates at 500 Hz. Hardware evaluation comprises fixed replay of highly dynamic AMASS motions and online Xsens teleoperation of highly dynamic and generalist motions. The AMASS references include Webster-style flip, butterfly kick, twisting back-handspring, and aerial cartwheel. The highly dynamic Xsens motions include standing tucked flip, air twist kick, running side flip, and air body twist, whereas the generalist set includes walking, calisthenic motions, crawling, and jumping. Each setting contains four representative motions, with five independent trials per motion.

\begin{table}[t]
\caption{Component-wise real-world success rates (\%). AMASS Replay uses fixed highly dynamic references, whereas Xsens Teleop. uses online Xsens inputs. $\dagger$ and $\ddagger$ denote highly dynamic and generalist Xsens motion evaluations, respectively. Each result is computed from four representative motions with five trials per motion.}
\label{tab:real_world_component_ablation}
\centering
\footnotesize
\renewcommand{\arraystretch}{1.15}
\begin{tabular*}{\columnwidth}{@{\extracolsep{\fill}}lcccc}
\toprule
Evaluation & Unified Enc. & w/o FSQ & w/o STAR & Ours \\
\midrule
AMASS Replay & 85.0 & 80.0 & 80.0 & \textbf{90.0} \\
Xsens Teleop.$^\dagger$ & 75.0 & 65.0 & 45.0 & \textbf{85.0} \\
Xsens Teleop.$^\ddagger$ & 90.0 & \textbf{100.0} & \textbf{100.0} & \textbf{100.0} \\
\bottomrule
\end{tabular*}
\end{table}

Building on these deployment results, Table~\ref{tab:real_world_component_ablation} quantitatively compares hardware performance and examines the contributions of the main design choices. Unified Enc. replaces the separate encoding branches and normalization interfaces for proprioceptive and action histories with a shared encoder. The w/o FSQ variant removes the finite scalar quantization constraint on the aggregated command representation, whereas w/o STAR removes the resampling of high-advantage trajectory fragments from difficult temporal regions. The complete method achieves the strongest performance in fixed replay and highly dynamic teleoperation while retaining stable control of generalist teleoperation motions.

All three variants reduce hardware performance. The degradation of Unified Enc. across the evaluations indicates that separately processing proprioceptive state and recent control responses reduces feature interference between these distinct histories. Removing FSQ makes online highly dynamic tracking more sensitive to rapid state variations and reference errors. The w/o STAR variant shows the largest degradation, particularly for highly dynamic Xsens teleoperation, which highlights the importance of prioritizing effective trajectory experience when inertial motion-capture inputs contain noise, root drift, and local pose errors.

Fig.~\ref{fig:replay} presents fixed-reference replay of representative highly dynamic motions, including aerial cartwheel, standing backflip, kip-up, and aerial twist. The robot completes aerial posture adjustment, rapid contact transitions, and landing recovery across these motions, demonstrating the physical executability of the learned policy for complex highly dynamic whole-body behaviors.

Fig.~\ref{fig:real_world_high_dynamic} further presents online Xsens teleoperation of standing tucked flip, air twist kick, running side flip, and air body twist. The policy constructs its reference window from the continuously arriving inertial motion-capture stream and therefore does not require a complete predefined trajectory. Unlike fixed-reference replay, the online inputs contain timing deviations, root drift, and local pose inconsistencies, while the amplitude, speed, and timing of operator motions vary naturally across trials. Under these more demanding conditions, the policy continuously responds to human inputs while performing rapid whole-body coordination, aerial posture adjustment, and contact transitions. These results demonstrate robustness to imperfect real-time references and the ability to directly execute diverse highly dynamic motions that are absent from the training corpus. The supplementary video provides additional demonstrations of Extreme-RGMT executing a broader range of highly dynamic and generalist motions.

\subsection{Limitations}

Extreme-RGMT tracks root-relative references and does not explicitly incorporate global position or heading information into the policy input. During long-duration execution, small tracking and state-estimation deviations can accumulate, causing drift in global orientation and position. Integrating global localization or periodic reference alignment may help reduce this drift in extended deployments.

Similar to human motor learning, acquiring highly dynamic motions may require dedicated practice. Accordingly, the policy has limited generalization to motions that differ substantially from the training distribution. Motions with unseen coordination patterns, timing, or contact configurations may require control behaviors that are insufficiently represented by the current training data. Extending the coverage of highly dynamic motion data and incorporating mechanisms for online adaptation are important directions for improving this capability.

\section{Conclusion}
\label{sec:conclusion}

This work presents Extreme-RGMT, which improves stable execution of highly dynamic motions for humanoid robots while retaining broad motion-tracking capability. The method learns a generalist base policy through an improved policy architecture, then combines PACE and STAR during the second stage to constrain base-policy drift and strengthen learning from high-advantage trajectory fragments.

Simulation results show that Extreme-RGMT achieves stable generalist joint-level tracking performance on both in-source and unseen motions, with coverage extended to highly dynamic motions. PACE supports specialist skill acquisition while retaining established capabilities, and STAR is particularly effective for highly dynamic motions captured by inertial motion-capture systems. Hardware experiments further validate execution under fixed highly dynamic references and online Xsens inputs, including direct tracking of diverse highly dynamic motions absent from the training corpus. Extreme-RGMT provides an effective approach to broadening generalist whole-body motion tracking toward highly dynamic motion regimes and advances generalist whole-body motion-tracking controllers toward highly dynamic motor capabilities at the human-expert level.

\bibliographystyle{IEEEtran}
\bibliography{main}

\end{document}